\documentclass[10pt,twocolumn,letterpaper]{article}

\usepackage{cvpr}              % To produce the CAMERA-READY version
\definecolor{cvprblue}{rgb}{0.21,0.49,0.74}
\usepackage[pagebackref,breaklinks,colorlinks,allcolors=cvprblue]{hyperref}
\usepackage{multirow}
\usepackage{booktabs}
\usepackage{makecell}
\usepackage{amsmath}
\usepackage{fontawesome}
\def\paperID{42881} % *** Enter the Paper ID here
\def\confName{CVPR}
\def\confYear{2026}

\title{TDATR: Improving End-to-End Table Recognition via \\Table Detail-Aware Learning and Cell-Level Visual Alignment}

\author{
Chunxia Qin\textsuperscript{\rm 1}\thanks{Equal contribution}\quad 
Chenyu Liu\textsuperscript{\rm 1,2}\footnotemark[1]\quad 
Pengcheng Xia\textsuperscript{\rm 2}\quad 
Jun Du\textsuperscript{\rm 1}\thanks{Corresponding author, jundu@ustc.edu.cn}\quad 
Baocai Yin\textsuperscript{\rm 2}\quad Bing Yin\textsuperscript{\rm 2}\quad Cong Liu\textsuperscript{\rm 2}\\
\textsuperscript{\rm 1}University of Science and Technology of China\quad \textsuperscript{\rm 2}iFLYTEK Research \\ 
{\tt\small cxqin@mail.ustc.edu.cn \quad Project Page: \href{https://github.com/Chunchunwumu/TDATR.git}{github.com/Chunchunwumu/TDATR.git}}
}

\begin{document}
\maketitle
\begin{abstract}
Tables are pervasive in diverse documents, making table recognition (TR) a fundamental task in document analysis. 
Existing modular TR pipelines separately model table structure and content, leading to suboptimal integration and complex workflows.
End-to-end approaches rely heavily on large-scale TR data and struggle in data-constrained scenarios.
To address these issues, we propose TDATR (Table Detail-Aware Table Recognition) improves end-to-end TR through table detail-aware learning and cell-level visual alignment.
TDATR adopts a “perceive-then-fuse” strategy. The model first performs table detail-aware learning to jointly perceive table structure and content through multiple structure understanding and content recognition tasks designed under a language modeling paradigm. These tasks can naturally leverage document data from diverse scenarios to enhance model robustness.
The model then integrates implicit table details to generate structured HTML outputs, enabling more efficient TR modeling when trained with limited data.
Furthermore, we design a structure-guided cell localization module integrated into the end-to-end TR framework, which efficiently locates cell and strengthens vision–language alignment. It enhances the interpretability and accuracy of TR.
We achieve state-of-the-art or highly competitive performance on seven benchmarks without dataset-specific fine-tuning.
\end{abstract}
 % This training strategy effectively decouples the core TR capabilities, reducing the difficulty of learning TR sequences from scratch.    
\section{Introduction}
\label{sec:intro}

\begin{figure}[t]
  \centering
   \includegraphics[width=1\linewidth]{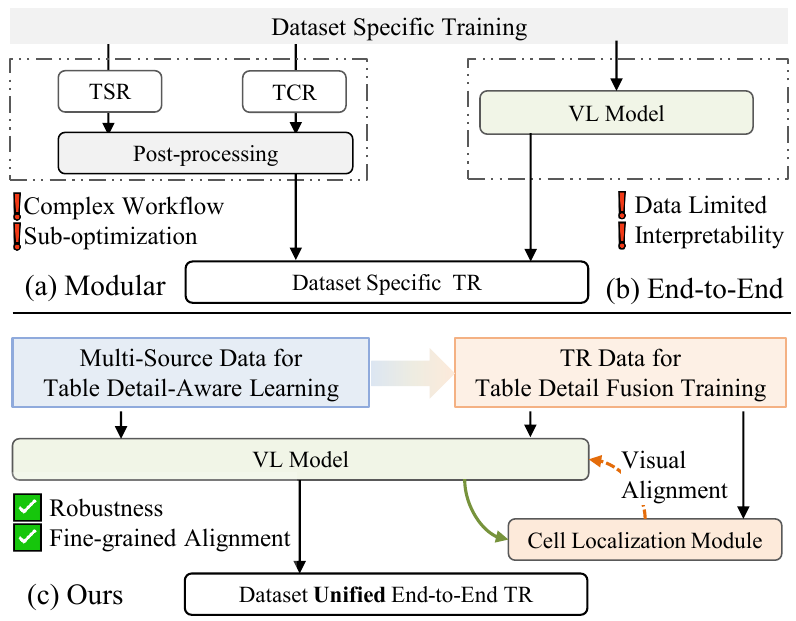}
   \caption{Comparison of different table recognition paradigms.
(a) Modular TR pipelines suffer from complex workflows and sub-optimization.
(b) End-to-end TR models underperform in data-scarce scenarios due to weak detail perception.
(c) Our ``perceive-then-fuse'' framework enhances structure and content awareness and unifies TR and cell localization for robust end-to-end TR.}
   \label{fig:intro}
\end{figure}

% Tables present content and the relationships between different content elements in a concise manner~\cite{Structural1993Chandran} through their specialized structures.
% Tables are widely used in documents across various domains~\cite{zhong2020pubtabnet,zheng2021gte}. With the advancement of computer technology, structured tables can be processed and analyzed automatically, as exemplified by tasks such as retrieval-augmented generation and document understanding. 
% A large number of unstructured tables in documents urgently need to be converted into machine-readable formats. 
% Table recognition (TR)~\cite{zanibbi2004survey} aims to transform unstructured inputs (e.g., table images) into structured outputs, such as HTML~\cite{zhong2020pubtabnet} and LaTeX~\cite{chi2019scitsr}. 

Tables convey structured data that bridges visual layouts and semantic information~\cite{Structural1993Chandran}.
Tables are pervasive across diverse domains such as scientific publications~\cite{zhong2020pubtabnet,chi_scitsr_2019}, invoices~\cite{long2021wtw}, and financial reports~\cite{zheng2021gte}. Table recognition (TR) converts table images into machine-readable formats (e.g., HTML, LaTeX).
Accurate TR facilitates downstream applications such as retrieval augmented generation~\cite{yu_tablerag_2025}, document understanding and document digitization~\cite{lv2023kosmos,blecher2023nougat}.

% However, end-to-end table recognition remains challenging due to issues such as low-quality table images, highly diverse table structures, difficulties in content recognition, and lack of data. These challenges underscore the critical need for more robust table recognition methodologies.

% Most table recognition studies decompose the task into two sub-tasks: table structure recognition (TSR)~\cite{Structural1993Chandran} and table content recognition (TCR)~\cite{shi2016end}. These studies focus on developing separate models for each sub-task, and the final table recognition result is obtained by fusing the individual outputs through post-processing~\cite{Anand2023tcocr}. 
Most existing TR systems follow a modular design, decomposing TR into two subtasks: table structure recognition (TSR)~\cite{Structural1993Chandran} and table content recognition (TCR)~\cite{shi2016end}.
Each component is trained independently, and the final TR result is obtained through post-processing~\cite{Anand2023tcocr}, as shown in ~\ref{fig:intro}(a).
However, this separation overlooks the inherent interdependence between structure and content, resulting in suboptimal integration and error accumulation. 
Specifically, table structures provide strong priors for constraining content boundaries, which guide the localization of text lines and prevent confusion between column separators and inter-character spacing.
Semantic continuity of cell content~\cite{zhang-etal-2024-unitabnet} helps distinguish visually adjacent cells, providing cues for structure recognition.

Recent studies~\cite{wan2024omniparser,zhong2020pubtabnet,feng_dolphin_2025} attempt to unify TSR and TCR into a single vision–language model that directly generates structured outputs.
While simplifying the pipeline, this paradigm relies heavily on large-scale annotated TR. Real world TR data are scarce due to the high cost of labeling both table structure and content~\cite{chen2022Matrix-Based-Augmentation}.
Consequently, end-to-end TR~\cite{zhao2024tabpedia,feng_dolphin_2025,zhong2020pubtabnet} models often struggle to generalize robustly to diverse real-world tables.
Moreover, most existing approaches only predict TR result without explicit spatial correspondence (e.g., cell locations)~\cite{zhao2024tabpedia,feng_dolphin_2025,zhong2020pubtabnet}, limiting the interpretability and applicability of TR results. 
Current end-to-end TR models typically rely on generic document~\cite{feng_dolphin_2025,Luo2024docowl1.5} or vision pre-training~\cite{peng2024unitable}, neglecting the fine-grained perception of table structure and content that is essential for precise table recognition.

To address these limitations, we propose  TDATR (Table Detail-Aware Table Recognition), a framework that enhances end-to-end TR through detail-aware learning and cell-level visual alignment.
TDATR follows a ``perceive-then-fuse'' strategy. In the perception stage, the model performs table detail-aware learning through our unified structure understanding and content recognition tasks under a language modeling paradigm. This equips model with strong table-detail perception and allows effective pre-training on large-scale and multi-domain data.
In the fusion stage, the model integrates the implicitly learned table details to generate structured HTML outputs using only limited TR data. 
This paradigm effectively decouples TR capability learning and alleviates the difficulty of modeling TR sequences from scratch.
Furthermore, we introduce a structure-guided cell localization module that efficiently localizes cell positions and strengthens vision–language alignment through structure priors and multi-level visual features, improving both interpretability and accuracy.
Experimental results on seven public benchmarks across different scenarios demonstrate the effectiveness and robustness of our method. Additionally, ablation studies further validate the efficacy of our key designs. 

Our main contributions are summarized as follows.
\begin{enumerate}
\item We propose a ``perceive-then-fuse'' strategy that reduces reliance on large-scale labeled TR data and simplifies the end-to-end sequence modeling of TR.
\item We design table detail-aware learning  that unifies structure understanding and content recognition through a set of pretraining tasks under a language modeling paradigm, enabling effective utilization of diverse document data to enhance model robustness.
\item We develop a structure-guided cell localization module that refines cell boxes via structure priors and multi-level visual features, enhancing visual alignment and TR accuracy.
\item We evaluate our unified model on seven public benchmarks without dataset-specific fine-tuning, demonstrating strong performance and robustness across diverse table styles and scenarios.
\end{enumerate}
\section{Related Work}
\label{sec:related_work}
% \subsection{Modular Table Recognition}
% Modular table recognition (TR) methods decompose the task into table structure recognition (TSR) and table content recognition (TCR). 
% The TSR model aims to infer the physical and logical coordinates of cells, while the TCR model focuses on text localization and recognition. 
% Under the split-and-merge paradigm, TSR models~\cite{zhang2022sem,ma2023RobusTabNet,beak2023TRACE,zhang2023semv2,wang2023tsrformer2,lyu2023gridformer} recover the grid structure by detecting row and column separators, then merging them into cells. 
% However, this approach assumes continuous boundaries for cells within the same row or column, making it difficult to handle misaligned tables. 
% Detection-based TSR methods~\cite{xing2023lore,liu2021flag,liu2022NCGM} detect individual cells and assign logical coordinates, but are limited by ambiguous boundaries in borderless tables.
% Image-to-markup based methods~\cite{zhang-etal-2024-unitabnet,Huang2023VAST} represent table structures as markup sequences (e.g., HTML or LaTeX) but require large-scale annotated data. 
% TCR models~\cite{Du2022SVTR,ye2023deepsolo,liu2021abcnet,peng2022spts,shi2016end} recognize text lines in tables and assign them to cells via IoU-based~\cite{Anand2023tcocr} or logical-based~\cite{peng2024unitable} matching. 
% However, since TSR and TCR are optimized independently, their inter-dependencies are underutilized, leading to error accumulation during fusion and suboptimal holistic understanding.
\textbf{Modular table recognition} methods employ two separate models for table structure recognition (TSR) and table content recognition (TCR). 
The TSR model aims to acquire the physical and logical coordinates of cells.
TSR models~\cite{zhang2022sem,ma2023RobusTabNet,beak2023TRACE,zhang2023semv2,wang2023tsrformer2,lyu2023gridformer} under the split-and-merge paradigm recover the grid structure of tables by detecting row and column separators, then merge grids into cells to generate TSR results. However, this approach assumes continuous boundaries for cells in the same row or column, making it difficult to handle misaligned tables. 
Detect-based TSR models~\cite{xing2023lore,liu2021flag,liu2022NCGM} obtain table structures by first detecting table cells and then recognizing their logical coordinates, but they are limited by ambiguous cell boundary definitions in borderless tables.
Image-to-markup based TSR methods~\cite{zhang-etal-2024-unitabnet,Huang2023VAST} represent table structures as markup sequences (e.g., HTML, LaTeX). However, they require large-scale training data.
The TCR models localize and recognize text lines in tables using existing OCR models~\cite{Du2022SVTR,ye2023deepsolo,liu2021abcnet,peng2022spts,shi2016end}.
Text lines are assigned to cells via IoU-based~\cite{Anand2023tcocr}  or logical-based~\cite{peng2024unitable} post-processing to produce final results.
However, since TSR and TCR are trained independently, their inter-dependencies cannot be exploited, leading to suboptimal performance and inevitable error accumulation in fusion.

\textbf{End-to-end table recognition} methods integrate TSR and TCR into a unified framework, and can be broadly classified into multi-decoder and single-decoder paradigms.
Multi-decoder methods adopt separate decoders for structure and content generation to decouple the modeling complexity of long TR sequences.
EDD~\cite{zhong2020pubtabnet} uses two decoders to decode structure tokens and cell content separately. 
Nam Tuan Ly et al.~\cite{ly_end--end_2023} propose an image encoder with three decoders, which generate table structure tokens, cell content, and cell boxes respectively.
OmniParser~\cite{wan2024omniparser} first generates a Structured Points Sequence to represent table structures and cell center coordinates, then uses these points as prompts to parse cell content.
Single-decoder methods utilize a unified decoder to decode table markup sequences. 
Dolphin~\cite{feng_dolphin_2025} models TR as HTML sequence.
To improve efficiency and eliminate redundancy in HTML representations, mPLUG-DocOwl1.5~\cite{hu2024mplug} adopts a concise Markdown-like format, while SmolDocling~\cite{nassar2025smoldocling}, Miner-U2.5~\cite{niu_mineru25_2025} and PaddleOCR-VL~\cite{cui-paddleocr-vl-2025} represent tables using OTSL~\cite{lysak_otsl_2023}.
Due to the inherent difficulty of TR, acceptable end-to-end TR performance in practice is often achieved by integrating expert OCR VLMs~\cite{wan2024omniparser,cui-paddleocr-vl-2025,niu_mineru25_2025,Wei2025DeepSeekOCRCO,dots.ocr} that rely heavily on large-scale document pre-training and extensive table-specific fine-tuning. However, these models largely overlook explicit perception of table structures, resulting in suboptimal utilization of structural cues and limited overall recognition quality.

% \subsection{Cell Localization}
\textbf{Cell localization} is a fundamental step that bridges visual table layouts and structured representations.
% While traditional TR pipelines treat cell localization as an independent subtask within modular systems, recent studies have explored incorporating explicit cell localization into end-to-end TR frameworks to enhance model interpretability.
Early approaches employ general object detection architectures~\cite{carion2020dert,ren2017fastrcnn,zhou2019centernet} to detect individual cells~\cite{Smock2022pubtables1m} within modular systems.
However, their performance degrades in dense or borderless tables due to ambiguous cell boundaries and extreme aspect ratios.
Subsequent works embed cell localization within end-to-end TR frameworks by leveraging hidden states of table representation tokens to regress~\cite{ly_end--end_2023} or generate~\cite{zhang-etal-2024-unitabnet} cell boxes.
Nevertheless,~\cite{ly_end--end_2023,Huang2023VAST} essentially predicts bounding boxes for cell contents, neglecting empty cells. 
Coordinate generation methods~\cite{zhao2024tabpedia,hu2024mplug,lv2023kosmos} represent cell positions with discrete tokens and generate them sequentially, which results in low efficiency for large tables.
To address these limitations, we design a structure-guided cell localization module, which fully exploits multi-level image features and structural priors to refine cell boundaries in parallel, achieving both higher accuracy and efficiency.

\section{Methodology}
\label{sec:method}

\begin{figure*}[t]
  \centering
   \includegraphics[width=1\linewidth]{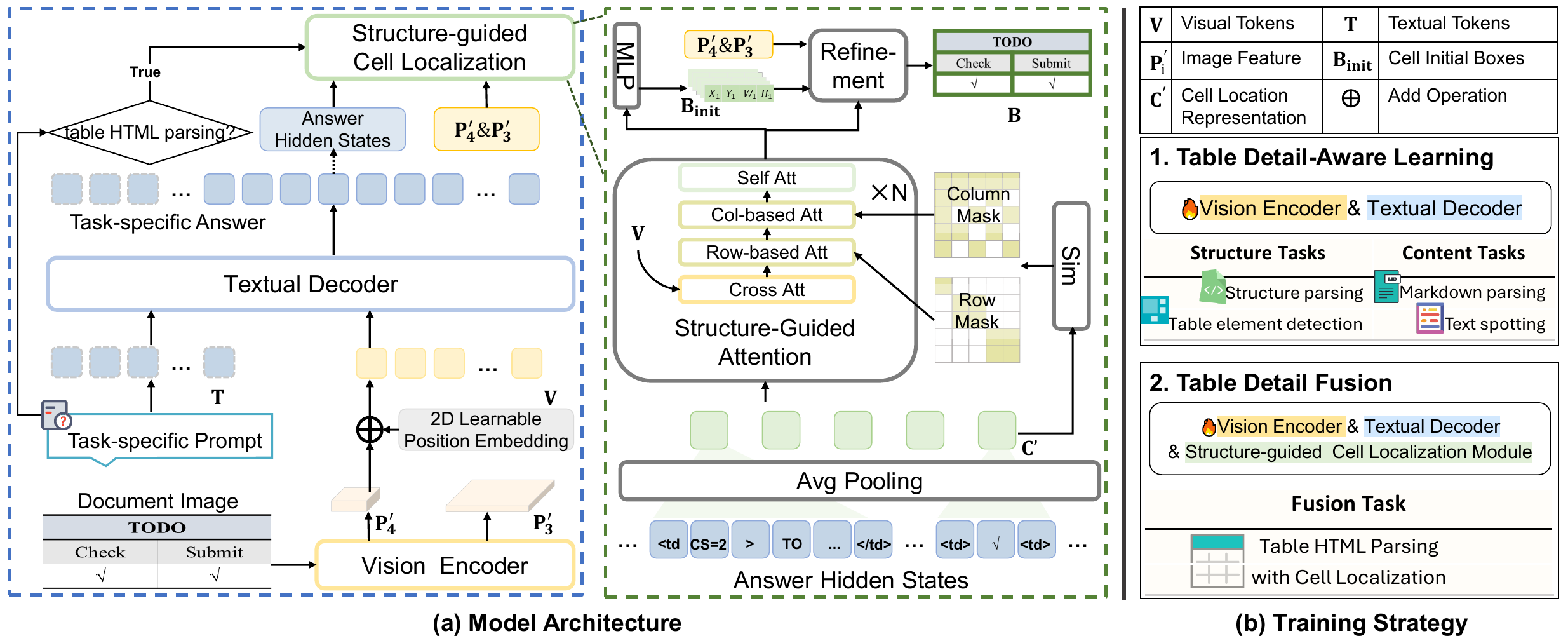}
   \caption{(a) The architecture of the model. The model consists of a d vision encoder, a language decoder, and a structure-guided cell localization module, which aggregates cell representations based on TR priors refines cell boxes using multi-resolution visual features. (b) The perceive-then-fuse training strategy for end-to-end table recognition. In the table detail-aware learning phase, we design table structure understanding and content recognition tasks under a language modeling paradigm to enhance fine-grained perception. In the fusion phase, we fine-tune the model for table HTML parsing by aggregating the learned implicitly table details, while jointly training the cell localization module to strengthen cell-level visual alignment.}
   \label{fig:model}
\end{figure*}
As illustrated in Fig.~\ref{fig:model}(a), our model consists of a visual encoder, a multi-modal language decoder, and the structure-guided cell localization module. Our method follows a ``perceive-then-fuse'' strategy to achieve accurate and robust end-to-end TR, as shown in Fig.~\ref{fig:model}(b).
In the perception stage, we perform table detail-aware learning. The model is pretrained to capture fine-grained table structure and content, under a unified language modeling paradigm.
In the fusion stage, the model generates the final structured table outputs from the visual and textual features learned during the perception stage. 
% To enhance performance, we introduce a \textit{structure-guided cell localization module}, which predicts cell boxes and strengthens vision-language alignment in parallel. 
The following sections elaborate on the model architecture and training strategy.

\subsection{Model Architecture}
% \subsubsection{Vison-Language Model}
Our vision-language model extracts multi-scale visual features and generates task-specific answer.

\textbf{Vision Encoder}. Inspired by works such as Donut~\cite{kim2022donut} and Dolphin~\cite{feng_dolphin_2025}, we adopt Swin Transformer~\cite{liu2021Swin} as our visual encoder. 
It encodes the input image into a feature pyramid $P=\{\textbf{P}_i \in \mathbb{R}^{d_i \times \frac{H}{2^i} \times \frac{W}{2^i}} \,|\, i=3,4,5\}$, corresponding to down-sampling rates of $8\times$, $16\times$, and $32\times$, respectively.
To enhance image features, we fuse adjacent resolution features as follows:  
\begin{equation}
    \textbf{P}_i' = \text{Conv}_{i1}(\textbf{P}_i) + \text{Conv}_{i2}(\text{Up}_{2\times}(\textbf{P}_{i+1})),\quad i=3,4
\end{equation}
Here, Conv denotes a 2D convolution operation, and $\text{Up}_{2\times}$ denotes a 2× upsampling operation. The enhanced features $\textbf{P}'_i$ are fed into the structure-guided cell localization module to refine cell boundaries.
For $\textbf{P}_4'$ , 2D learnable absolute positional embeddings~\cite{Xu2020LayoutLMv2MP} are appended, and the result is flattened to obtain the visual tokens $\textbf{V}$, which are subsequently used to enrich textual representations within the language decoder.  

\textbf{Language Decoder}. 
Inspired by works such as Donut~\cite{kim2022donut}, we construct a language decoder by stacking $l_s$ causal self-attention blocks~\cite{Vaswani2017transformer} and $l_c$ cross-attention blocks~\cite{Lin2021crossatt} to model cross-modal interactions. A text embedding module is employed to embed task-specific prompts into textual tokens $\textbf{T}$. In the cross-attention blocks, visual tokens $\textbf{V}$ serve as keys and values, while textual tokens act as queries.
The textual decoder generates task-specific answers via next token prediction, following the textual tokens.
% \subsubsection{Structure-Guided Cell Localization Module}

\textbf{Structure-guided cell localization (SGCL)}.   The SGCL module leverages the hidden states of the language decoder to refine cell boxes.
We first extract cell-level representations from the hidden states $\textbf{h}_i$ of different layers and token positions in the language decoder.
Shallow layers capture more visual cues, while deeper layers encode linguistic and structural information~\cite{Wang2024MLLMCS}. 
We aggregate these hidden states of different layers using learnable weights $w_i$ to obtain $\textbf{H}$.
Next, for each cell, we perform average pooling over the range between between the ``\texttt{<td}'' and ``\texttt{</td>}'' tokens to obtain the initial cell representation $\textbf{C}$.
To exploit spatial correlations among cells within the same row or column, we project $\textbf{C}$ into row and column feature spaces via linear layers.
\begin{equation}
    \textbf{C}^{k} = \text{Linear}^{k}(\textbf{C}), \: k=\text{row},\,\text{column}
    \label{eq:linear}
\end{equation}
We compute adjacency matrices from pairwise inner products of cell representations and derive structure masks $\textbf{M}^k$ through thresholding. 
\begin{equation}
    \textbf{M}^k_{xy} = \left \{
    \begin{aligned}
          1, \quad&\text{Sigmoid}(\langle\textbf{C}^k_x, \textbf{C}^k_y\rangle / \text{dim}(\textbf{C}^k)) \,>\,0\\
          0, \quad& Others\\
    \end{aligned}\right.
    \label{eq:sim}
\end{equation}
The obtained masks are then used to guide bidirectional contextual attention, enhancing $\textbf{C}$ to obtain the enhanced representation $\textbf{C}'$, as illustrated in Fig.~\ref{fig:model}(a). A more detailed illustration of the row-based cell representation feature enhancement is provided in Appendix~\ref{sup:sgcl}.

We regress initial cell boxes $\textbf{B}_{\text{init}}$ based on $\textbf{C}'$ using a simple MLP. 
To mitigate overlaps and positional offsets caused by the language decoder’s bias toward linguistic features, we further refine $\textbf{B}_{\text{init}}$ with multi-resolution visual features $\textbf{P}'_3$ and $\textbf{P}'_4$ through $l_d$ DAB-DETR decoder layers, yielding accurate cell boxes $\textbf{B}$.
Unlike standard DAB-DETR~\cite{liu2022dabdetr}, our anchors are initialized from the hidden states of TR cell representations, ensuring a one-to-one correspondence with TR outputs and eliminating the need for post-processing.
We further remove the unstable bipartite matching process to stabilize training and accelerate convergence.

\subsection{Training Strategy}
End-to-end table recognition requires three essential capabilities: table structure understanding, table content recognition, and table detail fusion~\cite{Huang2023VAST,zhang-etal-2024-unitabnet}. 
While previous works often learn these abilities jointly from large-scale TR data, our approach follows a ``perceive-then-fuse'' paradigm. 
We first perform table detail-aware learning to establish structure and content perception. The model then learns table HTML parsing to aggregate implicitly learned table details, accomplishing table recognition with explicit cell localization.

\subsubsection{Table Detail-Aware Learning}
This stage aims to endow the model with both table structure understanding and table content recognition capabilities under a unified language modeling framework.

\textbf{Table content recognition.}
To develop content recognition, we design three multi-granularity OCR tasks inspired by Kosmos 2.5~\cite{lv2023kosmos}.
Leveraging large-scale rich-text corpora from diverse sources, these tasks equip the model with fundamental abilities in text recognition, text localization, and reading order comprehension.
The use of diverse visual-text data enhances the model’s robustness to complex documents and reduces its reliance on specialized table datasets.

\textit{Spatially ordered text spotting}. The model outputs text lines in their spatial reading order, with an optional coordinate-free variant focusing solely on content.
This task builds basic text recognition and localization capabilities.

\textit{Text spotting with box query}.
Given a document region specified by a bounding box, the model performs spatially ordered text spotting to recognize and localize text lines.
A coordinate-free variant focuses solely on textual extraction.
This task enhances the  localization capability of model.

\textit{Markdown parsing.}
The model converts document images into Markdown format, reconstructing both textual content and layout, thereby developing document layout awareness.

\textbf{Table structure understanding}. To enhance table structure understanding, we designed table structure understanding tasks. These tasks are divided into cell-level and row-column-level tasks, allowing the model to perceive table structures at multiple hierarchies.

\textit{Table element detection tasks}.
The table cell detection task outputs cell coordinates in logical order, enabling the model to perceive cell spatial extents.
The span cell detection task predicts the coordinates of span cells together with their corresponding row and column ranges.
Since span cells are a major challenge in table recognition, this specialized task is designed to enhance the model’s perception of hierarchical table structures and spatial dependencies among span cells.
The row and column detection task sequentially outputs row and column boundaries, followed by the corresponding cells within each.
Modeling rows and columns encourages global structural perception, while their alignment with cell boundaries enhances the model’s understanding of span relationships.

\textit{Table structure parsing}.
This task outputs the structural representation of a table (in Markdown or HTML format), enabling the model and perceive the global logical organization of table elements.

In table detail-aware learning, all tasks adhere to the next-token prediction paradigm, and are supervised by cross-entropy loss $L_\text{ce}$.

\subsubsection{ Table Detail Fusion Fine-tuning}
After detail-aware learning, the model gains strong awareness of structural and textual elements.
We then conduct fusion fine-tuning to integrate these details for end-to-end table recognition.
Specifically, the model is trained on an HTML-based table parsing task, where it directly generates HTML sequences that jointly encode table structure and content.
Meanwhile, the cell localization module is optimized to predict precise cell coordinates, ensuring spatial consistency with textual outputs.
% The two jointly yield a complete TR result that combines semantic and geometric representations.

% After table detail-aware learning, the model can effectively perceive table structures and cell content.
% We implicitly leverage this table detail information through a table HTML parsing task to fuse table details, directly generating TR outputs. HTML provides a representation of both table structure and content. Meanwhile, we train the CBPR module to parse cell location information. This complements the HTML-based structural representation with physical cell coordinates, forming a complete TR result.

In table detail fusion fine-tuning, the table HTML parsing task also conforms to the next-token prediction paradigm with cross-entropy loss $L_\text{ce}$. For the SGCL module, we design three types of losses. A regression loss $L_\text{b}$ and an IoU loss $L_\text{iou}$~\cite{liu2022dabdetr} for cell regression. A mask alignment loss $L_\text{m}$ using a Mask-DINO~\cite{Li_2023_CVPR_mask_dino}-style segmentation head to enhance the alignment between cell representation $\textbf{C}'$ and image features $\textbf{P}'_4$;
A structure-guided loss $L_\text{s}$ to optimize the cell row–column relationship matrix using BCE loss.
The final fine-tuning loss is denoted as $L_\text{f}$, where $\lambda_i$ represents the weight corresponding to each loss. In practical experiments, we adjust $\lambda_i$ to balance the magnitudes of all losses.
\begin{equation}
    L_\text{f} = \lambda_\text{ce}\times L_\text{ce} + \lambda_\text{b}\times L_\text{b}+ \lambda_\text{iou}\times L_\text{iou}+\lambda_\text{m}\times L_\text{m}+ \lambda_\text{s}\times L_\text{s}
\end{equation}
\section{Data Preparation}
\label{sec:data}
The data used in our experiments can be categorized into two main groups, document data and table data.
% 因为文本识别数据主要决定了模型对于文本内容和文本位置的感知能力，以及模型对于真实场景的鲁棒性，所以我们使用了更加丰富的数据。

\subsection{Document Data}
As table content recognition data primarily govern the model’s ability to understand textual content, spatial layouts, and robustness in real-world scenarios, we utilize a large and diverse collection of Chinese and English document datasets to ensure strong generalization across domains and scenarios.
We collected a substantial amount of Chinese and English data for content recognition, shown in Table.~\ref{tab:ocr_data}. All data are used in spatially ordered text spotting task and text spotting with box query task. README files are used in Markdown parsing task.

For data from different sources, we employed distinct processing workflows due to their varying formats~\cite{lv2023kosmos,blecher2023nougat}. More details are provided in Appendix~\ref{sup:Documentdata}.

\begin{table}
\centering
\caption{The data are used for table content recognition tasks. ``ZH'' and ``EN'' denote Chinese and English datasets, respectively. ``R'' indicates real-world data. ``D'' represents digitally-born data. }
\scalebox{0.78}{\begin{tabular}{llcl} 
\toprule
Data Source                                 & \multicolumn{1}{c}{Number}             & Samping Rate & Type  \\ 
\hline
Webpage                                     & ZH 2.1M, EN 12.3M  & 0.2           & R,D   \\
Paper                                       & ZH 71M,~ EN 55.6 M & 0.4           & D     \\
WuKong~\cite{gu2022wukong} & ZH 42.2M           & 0.1           & R     \\
README                                      & 1.1M               & 0.1           & R,D   \\
In-house                                    & 12M                & 0.2           & R,D   \\
\bottomrule
\end{tabular}}

\label{tab:ocr_data}
\end{table}
\begin{table}[t]
\centering
\caption{Table data statistics used in the table structure understanding tasks and table HTML parsing task. The amount of real-world table data is limited.}
\scalebox{0.82}{\begin{tabular}{lr|lr} 
\toprule
\multicolumn{2}{c|}{Real-World Tables} & \multicolumn{2}{c}{Digitally-Born Tables}  \\ 
\hline
Data Source & Number                   & Data Source      & Number                  \\ 
\hline
iFLYTAB~\cite{zhang2023semv2}     & ZH 12k                   & PubTables-1M~\cite{Smock2022pubtables1m}     & EN 721k                 \\
iFLYTAB-Aug & ZH 82K                   & PubTabNet~\cite{zhong2020pubtabnet}        & EN 489k                 \\
WTW~\cite{long2021wtw}         & 10K                      & Table generation & ZH 924k                 \\
TabRecSet~\cite{Yang2023tabrecset}   & 30.5K                    & Re-render table  & ZH 184k                 \\
\bottomrule
\end{tabular}}

\label{tab:table_data}
\end{table}

\subsection{Table Data}
The table data are collected from both public datasets and synthetic corpora, as summarized in Table~\ref{tab:table_data}.
We employ two complementary synthesis strategies: (1) table generation, which produces tables with complex layouts, and (2) re-rendering web-crawled HTML tables to introduce realistic structures and diversify the data distribution.
We further augment real-world table recognition data using an improved Identity Matrix-Based Augmentation~\cite{chen2022Matrix-Based-Augmentation}, which crops table sub-regions for enrichment.
Annotations from heterogeneous table datasets are unified into a consistent format to enable consistent data usage across all table-related tasks.
All table data are used for table detail-aware learning.
For table detail fusion fine-tuning, we sample data from five public datasets, including iFLYTAB-full~\cite{zhang2023semv2}, TabRecSet~\cite{Yang2023tabrecset}, PubTabNet, PubTables-1M~\cite{Smock2022pubtables1m}, and FinTabNet~\cite{zheng2021gte}, covering diverse table structures and languages.
Additional implementation details are provided in Appendix~\ref{sup:Tabledata}.

To establish a challenging benchmark for Chinese table recognition, we manually completed the text annotations in the iFLYTAB~\cite{zhang2023semv2} dataset, forming a new dataset termed \textbf{iFLYTAB-full}. 
This dataset, which will be released publicly, contains a variety of wireless and camera-captured tables with complex structures and degraded image quality, closely reflecting real-world scenarios.
\section{Experiment}
\label{sec:experiment}

\subsection{Implementation Details}
We adopt the Donut Chinese model with a Swin-Transformer (300M) as the visual encoder, and a Transformer-based decoder (300M) as the language decoder, consisting of $L_s = 6$ causal self-attention blocks and $L_c=3$ cross-attention blocks. In the structure-guided cell localization (SGCL) module, the bidirectional enhancement branch includes 2 self-attention blocks and 1 cross-attention block.
The number of DAB-DETR decoder layers $L_d$ is set to 3.
Both the visual encoder and the language decoder are jointly optimized in both training stages, while the structure-guided cell localization module is trained only during the fine-tuning stage.
After fine-tuning, we obtain a unified end-to-end table recognition model without performing any additional fine-tuning on individual datasets.
Each stage is trained for 3 epochs using 16×64GB 910B NPUs.
The maximum decoding length is set to 4096 tokens.
Input images are resized so that both width and height are multiples of 256, and the longer side does not exceed 2048 pixels.
All element locations are represented by rectangular bounding boxes, defined by the top-left and bottom-right corner coordinates, which are normalized to the image size.
In generative tasks, coordinates are discretized, while in the structure-guided cell localization module, they remain continuous to support precise regression. We balance the training objectives by weighting their losses with empirical coefficients,
$\lambda_{b}=0.05$, $\lambda_{iou}=0.03$, $\lambda_{m}=0.03$, $\lambda_{s}=0.05$, and $\lambda_{ce}=1.0$.
Additional implementation details are provided in Appendix~\ref{sup:details}.

\subsection{Evaluation Benchmarks and Metrics}
We evaluate our method on seven table recognition benchmarks. These benchmarks together span diverse domains, languages, and scene conditions, enabling a thorough evaluation of our model.
The iFLYTAB-full~\cite{zhang2023semv2} and TabRecSet~\cite{Yang2023tabrecset} datasets are derived from real-world Chinese and English scenarios, featuring challenging cases such as contain challenging cases such as borderless tables, table region deformations, and low image quality. 
PubTabNet~\cite{zhong2020pubtabnet} and PubTables-1M~\cite{Smock2022pubtables1m} consist of English digital tables sourced from the PMCOA corpus. Notably, tables in PubTables-1M exhibit higher structural consistency, effectively mitigating the over-segmentation ambiguity observed in PubTabNet. 
OmniDocBench 1.5~\cite{omnidocbench_2025_CVPR}, CC-OCR~\cite{yang2024ccocrcomprehensivechallengingocr}, and OCRBench v2~\cite{fu2024ocrbenchv2improvedbenchmark}, which are originally designed for evaluating the OCR performance of multimodal large models.
We retain only samples related to table recognition. Additional details about the evaluation benchmarks are provided in Appendix~\ref{sup:eval_bench}.

We evaluate the effectiveness of table recognition using Tree-Edit-Distance-based Similarity (TEDS)~\cite{zhong2020pubtabnet}.
For table structure recognition, we report TEDS-S(tructure).
To measure table content accuracy, we adopt TEDS-Delta, defined as: $\text{TEDS-Delta} = \text{TEDS} - \text{TEDS-S}$.
For cell detection, we use the $\text{AP}_{50}$ metric~\cite{lin_coco_2014}, considering all table cells, including borderless and empty cells.

\subsection{Table Recognition Results}
We compare our method with expert TSR models, modular TR systems (M-TR), end-to-end TR models (E2E-TR), and expert OCR VLMs across two dimensions, table recognition and table structure recognition.
These comparisons comprehensively validate the effectiveness of our approach from both structural and content perspectives.
Notably, a single unified model is evaluated on all benchmarks without any dataset-specific fine-tuning, demonstrating strong generalization and robustness.
We further provide qualitative HTML parsing results for various table types in Appendix~\ref{sup:tr_vis}.

\begin{table}
\centering
\caption{Comparison with state-of-the-art methods on TabRecSet and iFLYTAB-full.
``*'' represents our reproduced results, which are obtained by training from scratch using open-source code and configurations. Bold denotes the first performances. ``+'' indicates that TR results are obtained by post-processing.
``$\dag$'' denotes results from a unified model without dataset-specific fine-tuning.
}
\scalebox{0.78}{
\begin{tabular}{llcc} 
\toprule
\multicolumn{4}{c}{\textbf{TabRecSet}}                                                                                  \\ 
\hline
\textbf{Type}   & \textbf{Method}                      & TEDS-S↑              & TEDS↑                 \\ 
\hline
\multirow{3}{*}{TSR}     & TableMaster~\cite{Ye2021TableMaster}                                   & 93.13                & -                     \\
                         & $\text{LORE}^{\ast}$~\cite{xing2023lore}                                         & 96.82                & -                     \\
                         & BGTR (PT)~\cite{hu-BGTR-2025}                                     & 97.21                & -                     \\ 
\hline
\multirow{2}{*}{E2E-TR}  & EDD~\cite{zhong2020pubtabnet}                                           & 90.68                & 70.70                 \\
                         & $\text{TDATR}^{\dag}$                                     & \textbf{ 97.27 }     & \textbf{ 92.70 }      \\ 
\toprule
\multicolumn{4}{c}{\textbf{iFLYTAB-full}}                                                                               \\ 
\hline
\multirow{3}{*}{TSR}     & $\text{LORE}^{\ast}$~\cite{xing2023lore}                                         & 87.83                & -                     \\
                         & UniTabNet~\cite{zhang-etal-2024-unitabnet}                                     & 94.00                & -                     \\
                         & BGTR (PT)~\cite{hu-BGTR-2025}                                     & 92.00                & {}  \\ 
\hline
M-TR                     & $\mathop{\text{SEMv3}^{\ast}}\limits_{\text{+PPOCR}}$~\cite{qin2024semv3} & 93.46                & 77.40                 \\ 
\hline
\multirow{3}{*}{OCR-VLM} & $\text{MinerU2.5}^{\dag}$~\cite{wang2024mineru}                                      & {64.16} & {58.47}  \\
                         & $\text{DeepSeek-OCR}^{\dag}$ ~\cite{Wei2025DeepSeekOCRCO}                                 & {77.44} & {84.36}  \\
                         & $\text{PaddleOCR-VL}^{\dag}$~\cite{cui-paddleocr-vl-2025}                                  & {76.04} & {81.48}  \\ 
\hline
E2E-TR                   & $\text{TDATR}^{\dag}$                                     & \textbf{96.59}       & \textbf{93.22}        \\
\cmidrule[\heavyrulewidth]{1-1}\cmidrule[\heavyrulewidth]{2-4}
\end{tabular}
}

\label{tab:real-world tab}
\end{table}
\textbf{Results on real-world tables}. 
As shown in Table~\ref{tab:real-world tab}, our method establishes new SOTA results on both iFLYTAB-full and TabRecSet. 
The TSR performance of our method  outperforms existing expert TSR models. This result highlights the beneficial impact of table content recognition on table structure recognition within end-to-end table recognition systems. 
Our method shows a significant performance gap compared to other TR approaches. Specifically, on iFLYTAB-full, it outperforms the modular TR method SEMv3+PPOCR by 15.82\% in TR performance. On TabRecSet, it surpasses the end-to-end TR method EDD by 6.59\% in TR performance. 
More importantly, our method requires far less fine-tuning data yet still achieves SOTA performance, highlighting its robustness and effectiveness.

\textbf{Results on digitally-born tables}. 
In digital scenarios, our method achieves SOTA TR performance on PubTabNet and PubTables-1M, as shown in Table~\ref{tab:pubtabnet}. Additionally, compared to modular TR methods, our approach achieves better alignment between table structure and content, mitigating error accumulation caused by post-processing. As demonstrated in Table~\ref{tab:pubtabnet}, our method achieves SOTA performance in TEDS-D.
Howere, our method slightly underperform the best TSR models. This gap mainly stems from two factors.
First, our fine-tuning uses only 0.4× of PubTables-1M and 0.6× of PubTabNet training data, resulting in a substantially limited data-fitting. In stark contrast to TableFormer~\cite{nassar2022tableFormer}, which relies on 24× more training data.
Further fine-tuning for two more epochs on PubTabNet (TDATR-ft) significantly improves both TSR and TR accuracy, confirming the benefit of additional data. 
Second, our method models complete TR sequences, which are about twice as long as TSR-only sequences, increasing generation difficulty. 
Nevertheless, our method outperforms Dolphin (a method with the same modeling approach) by 2.5\% in TEDS on both datasets, validating the effectiveness of our table detail aware learning. 

\begin{table}
\centering
\caption{The comparison result on Pubtables-1M and PubTabNet. ``*'' represents our reproduced results, which are obtained by inference using the released weights and official code. \underline{Underline} denotes the second-best performance. ``-ft'' denotes further fine-tuning of our unified model on the PubTabNet. ``PDF'' and ``GT'' denote table content extracted from the PDF source file and the ground-truth annotations, respectively.}
\setlength\tabcolsep{4pt} 
\scalebox{0.75}{
\begin{tabular}{llccc} 
\toprule
\multicolumn{5}{c}{\textbf{PubTables-1M}}                                                                                                                  \\ 
\hline
\textbf{\textbf{Type}}   & \textbf{\textbf{Method}}                                  & TEDS-S               & TEDS                 & TEDS-D↓               \\ 
\hline
\multirow{3}{*}{TSR}     & UniTabNet~\cite{zhang-etal-2024-unitabnet}                                                & \textbf{98.73}       & -                    & -                     \\
                         & $\text{TabPedia}^{\dag}$~\cite{zhao2024tabpedia}                               & 95.66                &                      &                       \\
                         & $\mathop{\text{DETR}}\limits_\text{+PDF}$                 & 97.65                & -                    & -                     \\ 
\hline
\multirow{2}{*}{OCR-VLM} & $\text{GOT}^{\dag}$~\cite{Wei2024GeneralOT}                                    & -                    & 36.84                & -                     \\
                         & $\text{Dolphin}^{\dag}$~\cite{feng_dolphin_2025}                               & 96.82                & \underline{95.48}        & 1.34                  \\ 
\hline
E2E-TR                   & $\text{TDATR}^{\dag}$                                   & \underline{98.39}        & \textbf{97.97}       & \textbf{0.42}         \\ 
\hline
\multicolumn{5}{c}{\textbf{PubTabNet-Val}}                                                                                                                 \\ 
\hline
\multirow{5}{*}{TSR}     & GTE~\cite{zheng2021gte}                                                       & 93.01                & -                    & -                     \\
                         & Davar-Lab~\cite{Ye2021TableMaster}                                                 & 96.36                & -                    & -                     \\
                         & $\text{TabPedia}^{\dag}$~\cite{zhao2024tabpedia}                               & 95.41                & -                    & -                     \\
                         & $\text{LORE}^{\ast}$~\cite{xing2023lore}                                                     & {94.55} & - & -  \\
                         % & TableVLM~\cite{chen_tablevlm_2023}                                                  & \textbf{96.92}                & -                    & -                     \\ 
\hline
\multirow{3}{*}{M-TR}    & $\mathop{\text{LGPMA}}\limits_\text{+R2AM}$~\cite{qiao2021lgpma}               & 96.70                 & 94.60                 & 2.10                   \\
                         & $\mathop{\text{TableFormer}}\limits_\text{+GT~~~~~~~~~~}$~\cite{nassar2022tableFormer} & \underline{96.75}        & 93.60                 & 3.15                  \\
                         & RapidTable~\cite{rapidtable}                                                & 96.43                & 86.57                & 9.86                  \\ 
\hline
\multirow{6}{*}{OCR-VLM} & $\text{DocOwl1.5}^{\dag}$~\cite{Luo2024docowl1.5}                             & 67.53                & 54.67                & 12.86                 \\
                         & $\text{OmniParser}^{\dag}$~\cite{wan2024omniparser}                             & 90.45                & 88.83                & 1.62                  \\
                         & $\text{Dolphin}^{\dag}$~\cite{feng_dolphin_2025}                                & 93.35                & 91.3                 & 2.05                  \\
                         & $\text{dots.ocr}^{\dag}$~\cite{dots.ocr}                               & 93.76                    & 90.65                    & 3.11                     \\
                         % & QwenVL 2.5-72B\dag                         & -                    & -                    & -                     \\
                         & $\text{MinerU 2.5}^{\dag}$~\cite{niu_mineru25_2025}                             & 93.11                     & 89.07                     &  4.04                     \\
                         & $\text{PaddleOCR-VL}^{\ast,\dag}$~\cite{cui-paddleocr-vl-2025}                          & 91.62                     & 87.27                      & 4.35                       \\ 
\hline
\multirow{3}{*}{E2E-TR}  & EDD~\cite{zhong2020pubtabnet}                                                       & 89.9                 & 88.3                 & 1.6                   \\
                         & $\text{TDATR}^{\dag}$                                   & 96.27                & 95.12                & 1.15                  \\
                         & TDATR-ft                                                    & \textbf{96.84}       & \textbf{96.10}       & \textbf{0.74}         \\
\bottomrule
\end{tabular}
}
\label{tab:pubtabnet}
\end{table}

\begin{table}
\centering
\caption{The comparison of our method with various MLLMs and expert OCR VLM for table recognition.}
\setlength\tabcolsep{3pt} 
\scalebox{0.7}{\begin{tabular}{lcccccc} 
\toprule
Method                                                       & \multicolumn{2}{c}{OmniDocBench1.5} & \multicolumn{2}{c}{CC-OCR}      & \multicolumn{2}{c}{OCRBenchv2}   \\ 
\cline{2-7}
                                                             & TEDS-S         & TEDS               & TEDS-S         & TEDS           & TEDS-S         & TEDS            \\ 
\hline
MiniCPM-V 4.5~\cite{Yao2024MiniCPMVAG}         & -              & -                  & 68.49          & 77.55          & 85.65          & 80.28           \\
InternVL3.5-241B~\cite{Wang2025InternVL35AO}  & -              & -                  & 62.87          & 69.52          & 85.81          & 79.50           \\
Qwen2.5-VL-72B~\cite{qwen2.5}               & -              & -                  & \underline{86.48}          & 81.22  & 86.58           &  81.33         \\
dots.ocr~\cite{dots.ocr}                    & 84.42          & 81.94              & 81.65          & 75.42          & 86.27          &  82.04          \\
MinerU2-VLM~\cite{wang2024mineru}           & 93.69          & 90.02              & 71.80          & 64.61           & 78.24          &  73.22          \\
MinerU2.5~\cite{niu_mineru25_2025}        & \underline{95.39}  & \underline{90.05}      & 85.16          & 79.76          & \underline{90.26}   & \underline{87.13}  \\
PaddleOCR-VL~\cite{cui-paddleocr-vl-2025} & \textbf{95.43} & \textbf{91.95}     & -              & -              & -              & -               \\
TDATR                                                         & 93.01          & 87.96              & \textbf{88.53} & \textbf{84.19} & \textbf{92.60} & \textbf{87.36}  \\
\bottomrule
\end{tabular}}

\label{tab:generic_model}
\end{table}

\textbf{Comparison with VLM on unseen domain}. 
We compared TDATR with general-purpose MLLMs, including MiniCPM-V 4.5~\cite{Yao2024MiniCPMVAG}, Qwen2.5-VL-72B~\cite{qwen2.5}, and QwenVL-2.5-72B~\cite{qwen2.5}, as well as expert OCR vision-language models (VLM), including dots.ocr~\cite{dots.ocr}, MinerU2-VLM~\cite{wang2024mineru}, MinerU2.5~\cite{niu_mineru25_2025}, and PaddleOCR-VL~\cite{cui-paddleocr-vl-2025}, for table recognition. As shown in Table~\ref{tab:generic_model}, our method achieves SOTA performance on CC-OCR and OCRBenchv2 among expert OCR-VLMs, and competitive performance compared to Gemini 2.5 Pro, while also performing strongly on OmniDocBench1.5. 
Notably, our model is substantially smaller and trained with far fewer resources. It requires only limited fine-tuning on publicly available datasets, whose scale and diversity are significantly lower than those used by other VLMs. These results demonstrate that TDATR generalizes effectively and exhibits strong robustness across diverse table scenarios. 

\subsection{Cell Localization Results}
\begin{figure}
    \centering
    \includegraphics[width=1\linewidth]{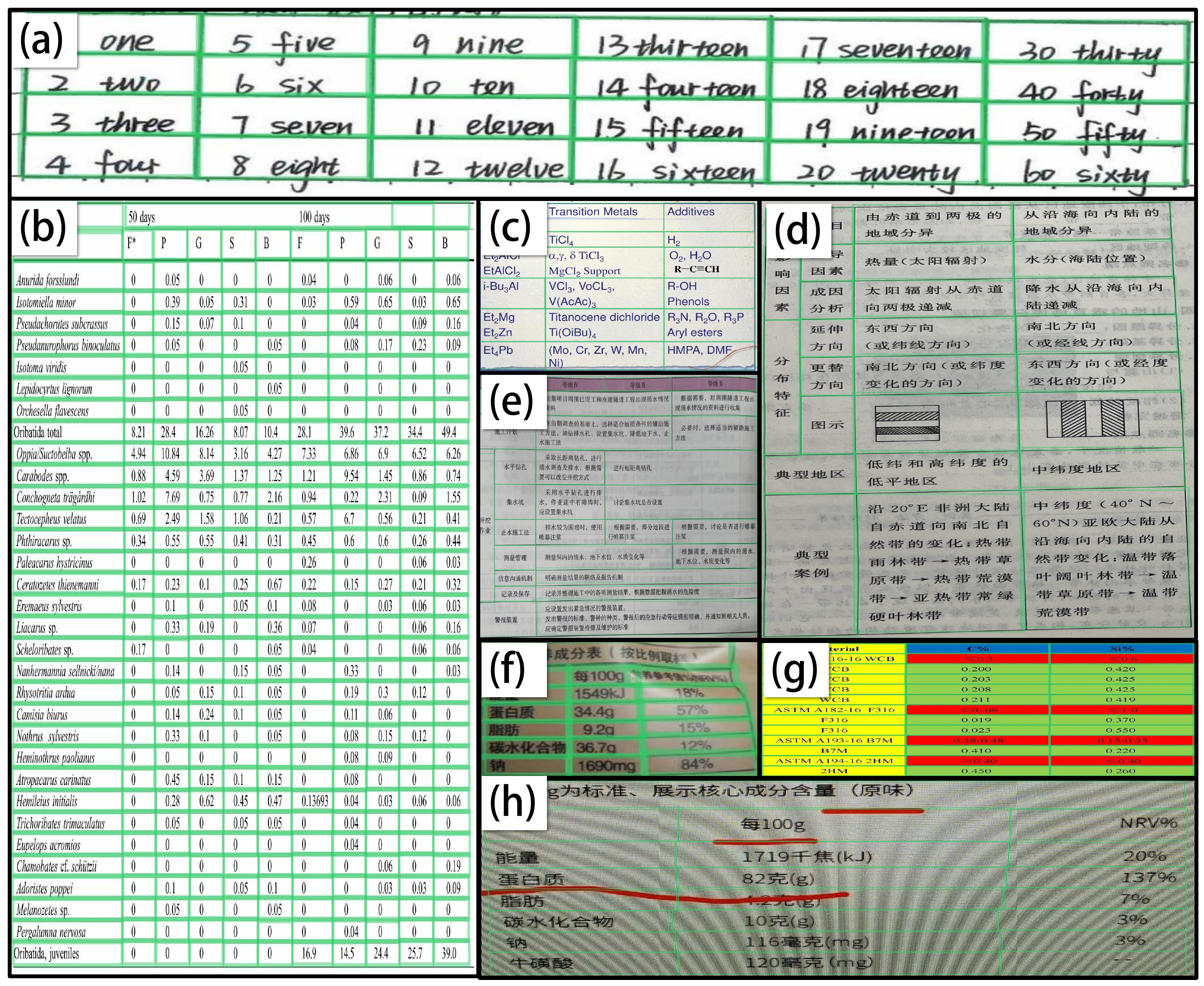}
    \caption{The visualization of cell localization on challenging tables, including borderless (b,h), complex-structured (e,d), long (b), and low-quality images (a,g,h).}
    \label{fig:vis}
\end{figure}

Table~\ref{tab:cell_loc} compares different cell localization methods. Our method achieves SOTA performance across both real-world and digital table scenarios.  
Visual-based methods (SEMv3, LORE) lack global table structure information, leading to ambiguous boundaries and suboptimal localization. UniTabNet leverages implicit cell information with location token classification, but compressing 8-point cell coordinates into a single token increases training difficulty. ED Loc Gen autoregressively generates interleaved TR HTML and cell locations, but at the cost of 33\% longer sequences and slower inference.  We further visualize cell localization results on representative challenging tables, including borderless, complex-structured, long, and low-quality images (see Fig.~\ref{fig:vis}).
In contrast, TDATR employs structure-guided parallel cell localization: implicit cell representations provide coarse localization, multi-resolution image features refine boundaries, and structure cues further enhance the representations. This design achieves more accurate localization, faster convergence, and efficient inference. Additional visual comparisons of cell localization are provided in Appendix~\ref{sup:cell_vis}.
\begin{table}
\centering
\caption{Comparison of table cell localization results with different localization methods on iFLYTAB-full and PubTabNet. ``ED Loc Gen'' denotes the using TDATR's encoder and decoder to autoregressively generate interleaved sequences of table recognition HTML and discrete cell coordinates. }
\setlength\tabcolsep{5pt} 
\scalebox{0.78}{
\begin{tabular}{llcc} 
\toprule
\multirow{2}{*}{Model} & \multirow{2}{*}{Cell Loc Method} & iFLYTAB-full & PubTabNet  \\ 
\cline{3-4}
                       &                                  & AP50         & AP50       \\ 
\hline
SEMv3*                  & Split-and-Merge                  & 92.92        & 85.12      \\
LORE*                   & CornerNet~\cite{Law2018CornerNetDO}                        & 91.87        & -          \\
UniTabNet*              & Loc
  Token Parallel Clf         & 88.43        & 89.67      \\
ED Loc Gen             & Loc
  Token Sequential Clf       & 93.52        & 90.26      \\
TDATR              & Structure-guided Cell Loc        & 94.37        & 91.80      \\
\bottomrule
\end{tabular}
}

\label{tab:cell_loc}
\end{table}

\subsection{Ablation Studies}
\textbf{The effectiveness of table detail-aware learning}.
Table detail fusion achieves end-to-end table recognition through table HTML parsing. We use table detail fusion as the baseline, as shown in Table~\ref{tab:ablation_tasks} T0. We explored the impact of able detail-aware learning on table recognition performance on the iFLYTAB-full and PubTabNet datasets, with results shown in the Table~\ref{tab:ablation_tasks}. Compared to the baseline T0, T1 and T2 demonstrate the contributions of table structure understanding tasks and table content recognition tasks to table recognition. T3 represents the complete two-stage training,  demonstrating the effectiveness of table detail-aware learning.
Table content recognition tasks contribute more significantly to table recognition, particularly on the iFLYTAB-full.
We attribute this to two factors: first, the large-scale and diverse document data enhances the robustness of the model. Second, it simultaneously improves table cell content recognition, improves text localization accuracy, and facilitates table structure restoration.

\begin{table}[t]
   \centering
  \caption{A Ablation study about table detail-aware learning (TDAL) on iFLYTAB-full and PubTabNet. "Content" denotes to table content tasks. And "Structure" refers to table structure understanding tasks. Table detail fusion (TDF) refers to the HTML parsing task conducted during the fine-tuning phase.}
   \setlength\tabcolsep{3pt}
   \scalebox{0.85}{
   \begin{tabular}{lccc|cccc} 
\toprule
   & \multicolumn{2}{c}{TDAL}                             & \multirow{2}{*}{TDF}    & \multicolumn{2}{c}{iFLYTAB-full} & \multicolumn{2}{c}{PubTabNet}  \\ 
\cline{2-3}\cline{5-8}
   & Content                   & Structure                 &                           & TEDS-S & TEDS                    & TEDS-S & TEDS                  \\ 
\hline
T0 &                           &                           & \checkmark & 89.29  & 82.63                   & 90.75  & 89.19                 \\
T1 &                           & \checkmark & \checkmark & 94.82  & 90.44                   & 94.30  & 92.45                 \\
T2 & \checkmark &                           & \checkmark & 95.02  & 91.57                   & 94.79  & 93.39                 \\
T3 & \checkmark & \checkmark & \checkmark & 96.11  & 92.50                   & 95.58  & 94.38                 \\
\bottomrule
\end{tabular}}

   \label{tab:ablation_tasks}
\end{table}

\begin{table}[t]
   \centering
    \caption{Ablation study on the structure-guided cell localization (SGCL) using iFLYTAB-full and PubTabNet. 
``Init-Reg'' refers to coordinate regression using cell representation features.
``Enh'' stands for the bidirectional attention cell enhancement.
``Ref'' indicates the cell coordinate refinement design. }
   \setlength\tabcolsep{4pt}
   \scalebox{0.8}{
\begin{tabular}{lcccc|cccc} 
\toprule
   & \multicolumn{4}{c|}{ SGCL }                                                                                     & \multicolumn{2}{c}{iFLYTAB-full} & \multicolumn{2}{c}{PubTabNet}  \\ 
\cline{2-9}
   & Init-Reg                  & Enh                       & Struct-M                  & Ref                       & TEDS  & $\text{AP}_{50}$      & TEDS  & $\text{AP}_{50}$         \\ 
\hline
T3 &                           &                           &                           &                           & 91.88 & -                   & 94.38 & -                      \\
C1 & \checkmark &                           &                           &                           & 93.42 & 87.44               & 94.78 & 89.63                  \\
C2 & \checkmark & \checkmark &                           &                           & 93.51 & 89.21               & 94.83 & 90.02                  \\
C3 & \checkmark & \checkmark & \checkmark &                           & 93.41 & 89.22               & 94.87 & 90.26                  \\
C4 & \checkmark & \checkmark & \checkmark & \checkmark & 93.52 & 94.37               & 94.80 & 91.81                  \\
\bottomrule
\end{tabular}
   }
  
   \label{tab:ablation_CBPR}
\end{table}

\textbf{The effectiveness of structure-guided cell localization}. As shown in Table \ref{tab:ablation_CBPR}, we demonstrate the structure-guided cell localization module. The C1-C4 designs can complement the cell position information for T3, expanding the application scenarios of TR model. Furthermore, C1-C4 integrate the HTML parsing task with cell positions, enhancing alignment between vision and language and improving TR performance.
Compared to C2, C3 introduces a structure mask in the bidirectional attention cell feature enhancement process, enabling cells to focus more on information from cells in the same row and column, thus improving cell detection performance. C4 further incorporates multi-resolution image features to refine cell regression results, achieving more accurate cell detection results. 
% As illustrated in the figure in Appendix~\ref{sup:cell_vis}, cell location results for different configurations are visualized. 
\section{Conclusion}
\label{sec:Conclusion}
In this work, we propose TADTR, an end-to-end framework that improves end-to-end TR through table detail-aware learning and cell-level visual alignment.  Through our “perceive-then-fuse’’ strategy, the model first acquires robust structural and textual awareness via table detail-aware learning, and then effectively transfers these capabilities to end-to-end TR with only limited supervised data. The proposed structure-guided cell localization module further enhances visual–structural alignment, enabling accurate cell-level spatial prediction while simultaneously improving the accuracy and interpretability of TR.
Extensive experiments on seven benchmarks demonstrate the superiority and robustness of our approach across diverse table types and layouts. 
Moreover, our framework is built upon the general VLM architecture, making its table detail-aware learning paradigm readily transferable to other document understanding and parsing VLMs. This provides a solution to improve the structural and textual perception of the table.

% In this work, we propose TADTR, an end-to-end framework that improves end-to-end TR through table detail-aware learning and cell-level visual alignment.  TADTR alleviates data dependency and modeling complexity through table detail-aware learning. Furthermore, the proposed structure-guided cell localization module leverages structural priors and multi-level visual features to refine cell boxes, which achieve precise spatial alignment and enhanced TR accuracy. Extensive experiments on iFLYTAB, TabRecSet, PubTabNet, PubTables-1M, OmniDocBench1.5, CC-OCR and OCRBenchv2 demonstrate the superiority and robustness of our approach across diverse table types and layouts. 

\section*{Acknowledgement}
This work was supported by the National Natural Science Foundation of China under Grant No. U25A20409. 
{
    \small
    \bibliographystyle{ieeenat_fullname}
    \bibliography{main}
}

% WARNING: do not forget to delete the supplementary pages from your submission 
\clearpage
\maketitlesupplementary
\appendix
% \setcounter{section}{0}
% 总体控制在6页以内

\section{Document Data Processing}
\label{sup:Documentdata}
For data from different sources, we employed distinct processing workflows due to their varying formats~\cite{lv2023kosmos,blecher2023nougat}.

Chinese and English webpages: We render HTML file to image using khtmltopdf\footnote{https://wkhtmltopdf.org/}. Then we utilized a commercial OCR\footnote{https://www.xfyun.cn/services/common-ocr} engine to recognize text lines on the webpages, extracting both the textual content and their corresponding coordinates. 

Chinese and English papers: For papers with LaTeX source code, we first compile the LaTeX code into a PDF, and then use the PyMuPDF\footnote{https://github.com/pymupdf/PyMuPDF} parser to extract text lines and their coordinates from the compiled PDF. For papers available only in PDF format, we utilize a commercial engine to extract text lines and coordinates. Specifically, for mathematical formulas in papers, we employ LatexOCR\footnote{https://github.com/lukas-blecher/LaTeX-OCR} tool.

README files: We downloaded README files and their referenced content from various GitHub projects. First, we filter out invisible elements from the README files, such as web links, jump markers, and comments, to ensure consistency between the text and rendered images. We then used Pandoc\footnote{https://pandoc.org/} to convert the filtered README files into HTML. Finally, we utilized wkhtmltopdf to convert the HTML content into images. To limit the image size, we segmented the images and extracted the corresponding markdown content as labels.

WuKong dataset~\cite{gu2022wukong} and in-house data: We utilized a commercial OCR engine to recognize text lines on images. 

\section{Table Data Processing}
\label{sup:Tabledata}
Real-world table refers to images captured through photographing or scanning. Such images often contain geometric distortions, background noise, and low resolution, making recognition considerably more challenging.
Digitally-born table refers to images  rendered directly from code or digital documents. These images have clean characters and well-aligned layouts.
\subsection{Unified Multi-source Table Data Processing}
To obtain labels for the table auxiliary tasks from various dataets, we designed a unified processing pipeline. 

In the first step, we unify the table label from different sources into a consistent format. In document images, we represent a table using table box, table cells, and table grids. The table box indicates the position of the table area within the document image. Table cells contain cell coordinates, logical coordinates, the text content within each cell and cell ID. Table grids represent the fine-grained structure of a table, showing the results after splitting merged cells. Each table grid includes the ID of the corresponding cell and its coordinates.

In the second step, we conduct data cleaning to eliminate inconsistently labeled table data, ensuring high data quality. First, we remove table data with overlapping logical coordinates for cells. Next, we exclude entries with incomplete table grids, specifically those where grids have not been assigned to their corresponding cells. Finally, we eliminate redundant table grids, which occur when adjacent grid rows and grid columns are identical.

The last step is training data generation. We extract table images from document images by cropping based on the table box. 
Table cell information is used for label generation in the table HTML parsing task, table cell detection task, and table cell spotting task. Table grid information is utilized for label generation in the table span cell detection task and the table row and column detection task.
\begin{figure}
    \centering
    \includegraphics[width=0.95\linewidth]{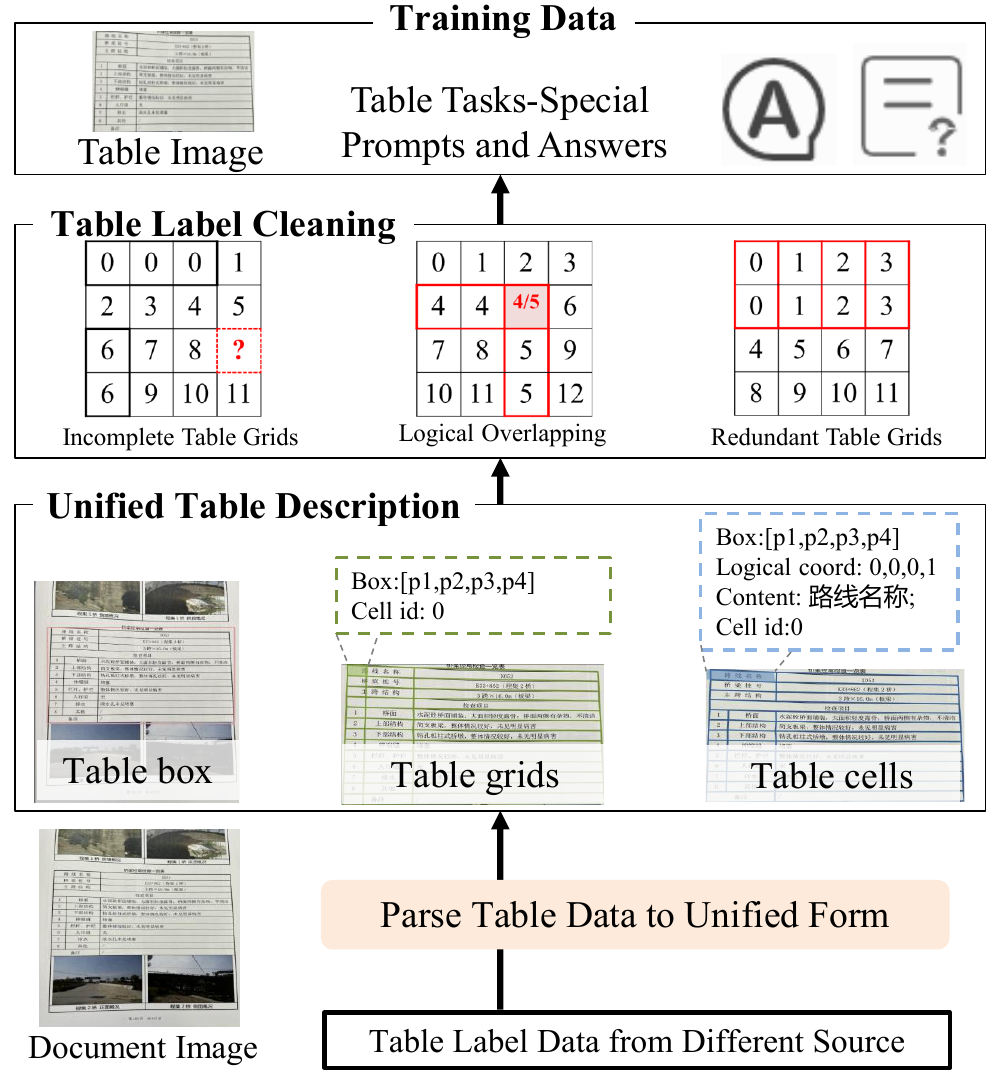}
    \caption{The pipeline of unified multi-source table data processing. The pipeline normalizes heterogeneous table annotations from various sources into a unified representation for model training.}
    \label{fig:placeholder}
\end{figure}

\subsection{Table Data Augmentation} 
High-quality labeled table data for photographic scenes is limited~\cite{zhang2023semv2,Yang2023tabrecset}. We expanded the iFLYTAB~\cite{zhang2023semv2} dataset inspired by an identity matrix-based augmentation method~\cite{chen2022Matrix-Based-Augmentation}, resulting in the iFLYTAB-Aug dataset with 82.5k samples. 

We made the following modifications to the identity matrix-based augmentation to ensure the generation of complex and realistic table data.
\begin{itemize}
    \item We restrict the selected table regions to have more than 4 rows and columns. 
    \item We ensure that the selected table sub-region always contains at least one span cell, and all rows and columns containing the span cell are retained. This ensures that the table has a complex structure.
    \item For wireless tables, the selected region always starts from the first row and the first column. Because the row and column headers provide essential information for distinguishing between rows and columns.
\end{itemize}

\section{Implementation Details}
\begin{figure*}
    \centering
\includegraphics[width=0.9\linewidth]{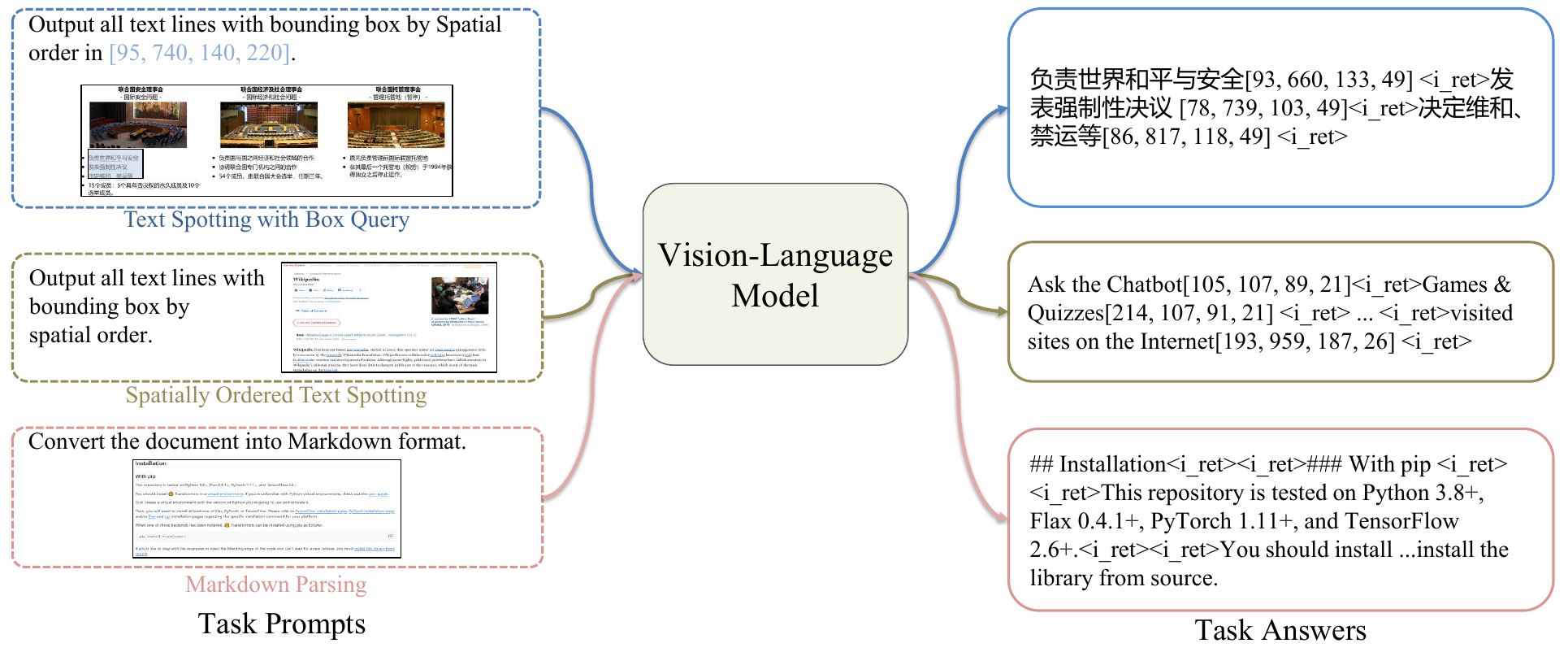}
    \caption{Illustration of table content recognition tasks. These tasks leverage diverse document data to enable text recognition, text localization, and reading-order understanding.}
    \label{fig:content-task}
\end{figure*}

\begin{figure*}
    \centering
    \includegraphics[width=0.9\linewidth]{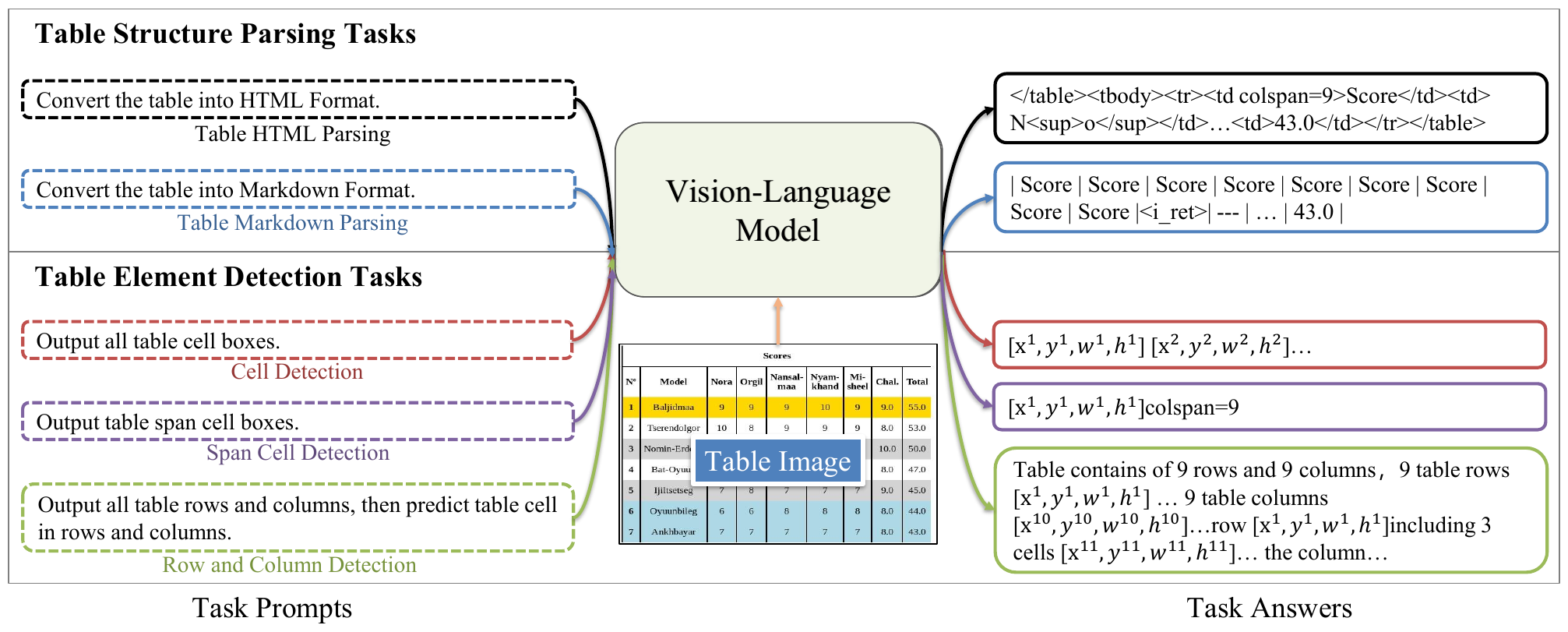}
    \caption{Illustration of table structure understanding tasks. These tasks equip the model with structure-awareness from both the cell level and the row/column level.}
    \label{fig:struct-tasks}
\end{figure*}

% \label{bench}
\label{sup:details}
In this section, we provide the detailed input–output designs of the table detail-aware learning tasks, as illustrated in the Fig.~\ref{fig:content-task} and~\ref{fig:struct-tasks}.

\section{Baseline Protocol}
\label{baseline_protocol}
Thanks for pointing out the ambiguity in Tables.~\ref{tab:real-world tab} and~\ref{tab:pubtabnet}. Our compared baselines can be grouped into:
(1) Dataset-specific setting: methods fine-tuned on each target dataset. 
(2) Unified setting (marked with “$\dag$”): a single checkpoint evaluated across multiple datasets. Table~\ref{tab:baseline_data_config} presents the training data configurations of all baseline methods used in this paper.
\begin{table*}[t]
\centering
\caption{Summary of the training data configurations of the baseline methods. For each method, we report the paradigm, table training data, auxiliary data, whether table-specific fine-tuning is applied, and additional notes.}
\scalebox{0.72}{
\begin{tabular}{llllll}
\toprule
Method           & Paradigm             & Table training data                                      & Extra data                                                                                                                                           & Dataset specific & Notes                                                                                                                                  \\ 
\midrule
TableMaster      & TSR                  & PubTabNet                                                & –                                                                                                                                                    & Yes                        & –                                                                                                                                      \\
LORE             & TSR                  & \makecell[l]{PubTabNet, TabRecSet,\\and iFLYTAB}                        & –                                                                                                                                                    & Yes                        & \makecell[l]{20k images were randomly sampled from \\PubTabNet for training. \\TabRecSet and iFLYTAB were reproduced \\by us based on the released code.}  \\
BGTR (PT)        & TSR                  & \makecell[l]{TabRecSet, iFLYTAB, \\PubTabNet, FinTabNet \\and SynthTabNet} & –                                                                                                                                                    & Yes                        & –                                                                                                                                      \\
UniTabNet        & TSR                  & \makecell[l]{iFLYTAB, PubTables-1M, \\and PubTabNet.}                   & \makecell[l]{Pre-training: a synthetic dataset \\comprising 1.4 million Chinese \\and English samples from \\SynthDog, and PubTables-1M} & Yes                        & –                                                                                                                                      \\
EDD              & E2E-TR               & PubTabNet                                                & –                                                                                                                                                    & Yes                        & –                                                                                                                                      \\
SEMv3 + PPOCR    & M-TR                 & PubTabNet and iFLYTAB                                    & \makecell[l]{PPOCR relies on general \\text recognition data }                                                                                                       & Yes                        & \makecell[l]{“+PPOCR” indicates that the cell content \\is obtained from the PPOCR model.}                                                             \\
GTE              & TSR                  & PubTabNet and FinTabNet                                  & –                                                                                                                                                    & Yes                        & \makecell[l]{The model is pre-trained on PubTabNet \\and fine-tuned on multiple datasets.}                                                             \\
Davar-Lab        & TSR                  & PubTabNet                                                & –                                                                                                                                                    & Yes                        & –                                                                                                                                      \\
LGPMA + R2AM     & M-TR                 & PubTabNet                                                & Additional data required by R2AM                                                                                                                     & Yes                        & \makecell[l]{“+R2AM” indicates that the cell \\content is obtained from the R2AM model.}                                                               \\
TableFormer + GT & M-TR                 & \makecell[l]{PubTabNet, FinTabNet, \\and SynthTabNet}                    & –                                                                                                                                                    & Yes                        & \makecell[l]{“+GT” indicates that the ground-truth cell \\content.}                                                                            \\
RapidTable       & M-TR                 & –                                                        & –                                                                                                                                                    & –                          & –                                                                                                                                      \\
OmniParser       & OCR-VLM & PubTabNet and FinTabNet                                  & Large-scale document parsing data                                                                                        & Yes                        & –                                                                                                                                      \\
DocOwl1.5        & OCR-VLM              & TURL and PubTabNet                                       & \makecell[l]{Unified structure-learning data \\from documents, webpages, charts, \\and natural images}                                                                 & No                         & –                                                                                                                                      \\
Dolphin          & OCR-VLM     & PubTabNet and PubTab1M                                   & Large-scale document parsing data                                                                                                                    & Yes                        & –                                                                                                                                      \\
MinerU2.5        & OCR-VLM              & In-house                                                 & Large-scale document parsing data                                                                                                                    & No                         & –                                                                                                                                      \\
DeepSeek-OCR     & OCR-VLM              & In-house                                                 & Large-scale document parsing data                                                                                                                    & No                         & –                                                                                                                                      \\
PaddleOCR-VL     & OCR-VLM              & In-house                                                 & Large-scale document parsing data                                                                                                                    & No                         & –                                                                                                                                      \\
dots.ocr         & OCR-VLM              & In-house                                                 & Large-scale document parsing data                                                                                                                    & No                         & –                                                                                                                                      \\
GOT              & OCR-VLM              & In-house                                                 & Large-scale document parsing data                                                                                                                & No                         & –                                                                                                                                      \\
TabPedia         & TSR                  & PubTabNet and PubTab1M                                   & –                                                                                                                                                    & No                         & –                                                                                                                                      \\
DETR + PDF       & M-TR                 & PubTables-1M                                             & –                                                                                                                                                    & Yes                        & \makecell[l]{“+PDF” indicates that the cell content \\is obtained from the source PDF files.}                                                          \\
\bottomrule
\end{tabular}}
\label{tab:baseline_data_config}

\end{table*}

\section{Structure-guided Cell Localization Module}
\label{sup:sgcl}
\begin{figure}
    \centering
    \includegraphics[width=0.95\linewidth]{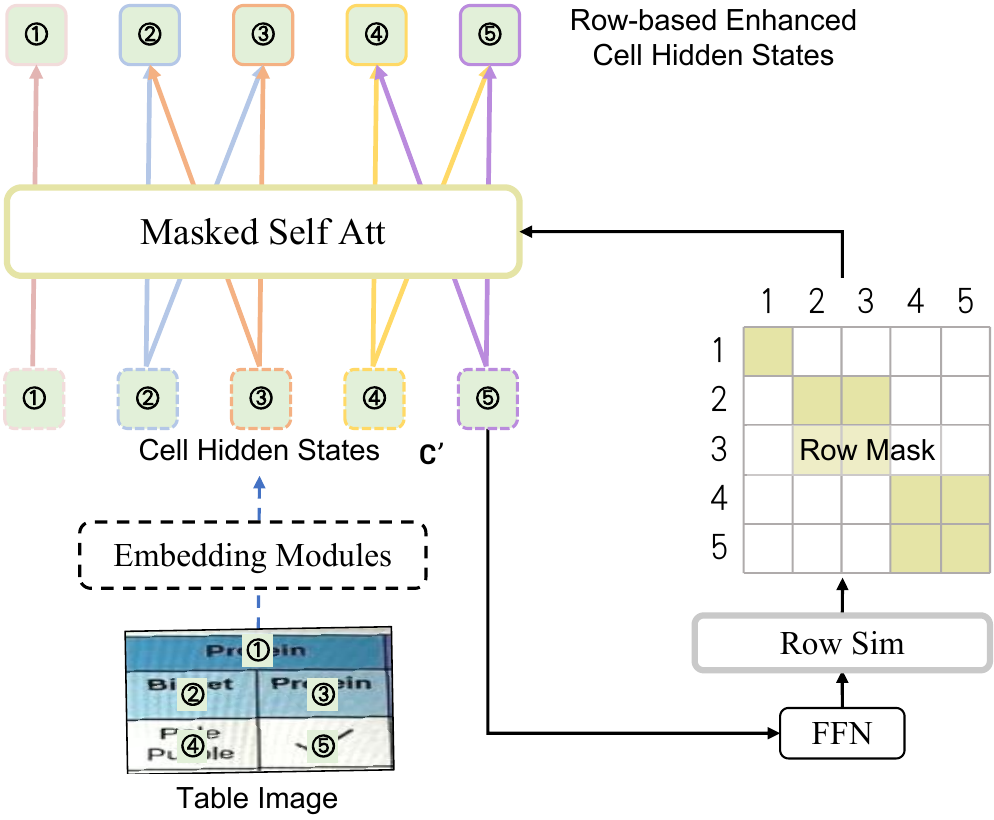}
    \caption{The architecture of the structure-guided cell localization module, illustrated with the row-based cell enhancement example.}
    \label{fig:sgcl}
\end{figure}

We leverage logical relationships between cells to perform bidirectional structure-guided enhancement. We take the row-based enhancement as an example to illustrate the computation process.
We first project the cell representation $\textbf{C}'$ into a row feature space using a linear layer to obtain row-level similarity features $\textbf{C}^{row}$, as shown in Eq.~\ref{eq:linear}. We then compute pairwise similarity scores via inner product to estimate whether two cells belong to the same row. After thresholding, we obtain the row similarity matrix, i.e., a binary relationship matrix indicating which cell pairs share the same row, as defined in Eq.~\ref{eq:sim}.
As illustrated in Fig.~\ref{fig:sgcl}, Cell~2 and Cell~3 are in the same row, thus $M^{row}_{2,3}=1$. This matrix is subsequently used as a mask in self-attention to reinforce feature interactions among cells within the same row.

\section{Evaluation Benchmarks}
\label{sup:eval_bench}
\textbf{iFLYTAB-full} obtains 5,419 test samples. The samples come from diverse sources—including screenshots, scans, and camera-captured images—covering a wide range of image qualities that allow evaluation of model robustness. The dataset exhibits large variations in image resolution, testing the model’s capability to handle multi-resolution inputs. It also contains grid tables, bordered three-line tables, and borderless tables. The absence of visible cell boundaries in borderless tables introduces significant challenges for TR.

\textbf{TabRecSet} contains 7,548 validation samples, all captured in real-world scenarios with strong perspective distortion and low image quality. Borderless and three-line tables are generated by erasing the ruling lines of grid tables, creating a domain gap between these synthetic styles and real-world data. The dataset includes both Chinese and English tables.

\textbf{PubTabNet} consists of 9,015 validation samples and 9,064 test samples, with the validation set commonly used for benchmarking. Its annotations are produced by an automated pipeline, resulting in low-resolution images and inconsistent visual-HTML alignment (e.g., cell over-segmentation). Such inconsistencies lead to contradictory training signals and may underestimate performance during evaluation. Models often require dataset-specific fine-tuning to adapt to these inconsistencies.

\textbf{PubTables-1M} contains 93,834 test samples and is sourced from the same corpus as PubTabNet. It applies automated consistency checks to correct the annotation inconsistencies present in PubTabNet, resulting in significantly improved label reliability.

\textbf{OmniDocBench v1.5}.
Following PaddleOCR-VL, we crop 512 table samples from the benchmark. The dataset covers a wide spectrum of table types, including challenging note-style tables where continuous content and background ruling lines visually disrupt cell boundaries, often causing over-segmentation. Successful recognition requires semantic understanding of cell content beyond visual boundary cues.

\textbf{CC-OCR}.
The 300 table test samples in CC-OCR cover both Chinese and English, spanning real-world and digital-document scenarios. The dataset includes long tables, dense tables, and heavily rotated cases, posing significant challenges for structure parsing and spatial reasoning.

\textbf{OCRBench} includes 700 table-related samples in both Chinese and English. Using the provided table boxes and our internal table detector, we crop table regions for recognition. Many samples come from financial reports, whose formatting introduces unique difficulties, e.g., large spacing between ``\$'' and numbers is easily mistaken for column separators.

\section{Single-dataset Training Variant}
\label{sup:single_dataset}
We conduct a single-dataset comparison by performing only-PubTabNet table detail fusion fine-tuning starting from our table detail-aware pretrained model. On PubTabNet-val, we achieve {TEDS-S 96.78 / TEDS 96.10}, outperforming the second-best TR dataset-specific baseline, TableFormer (96.75 / 93.60). This supports the effectiveness of our ``perceive-then-fuse'' paradigm in a single-dataset setting.

\section{SGCL Inference Efficiency  }
\label{sup:sgcl_speed}
TDATR leverages SGCL to localize cells in parallel, conditioned on the generated cell tokens. Since TR baselines such as Dolphin or EDD do not output cell boxes, a direct efficiency comparison is not applicable.
For a fair comparison, we implement a matched baseline, ``ED Loc Gen'' in Table.~\ref{tab:cell_loc}, that autoregressively generates discretized coordinates after the cell tokens. We evaluate 40 randomly sampled PubTabNet images (max side length 1024), with an average of 26.75 cells and 190.38 TR tokens.
Measured on an NPU with batch size 1, TDATR achieves 9.7s end-to-end latency, compared to 15.7s for the baseline (1.6$\times$ faster).  Importantly, SGCL contributes only 0.28s to the end-to-end latency, confirming that parallel refinement keeps localization overhead low. TDATR and the baselines have comparable max reserved memory (15.77\,GiB vs.\ 15.36\,GiB).

\section{Visualization of Table Recognition}
\label{sup:tr_vis}
\begin{figure*}
    \centering
    \includegraphics[width=1\linewidth]{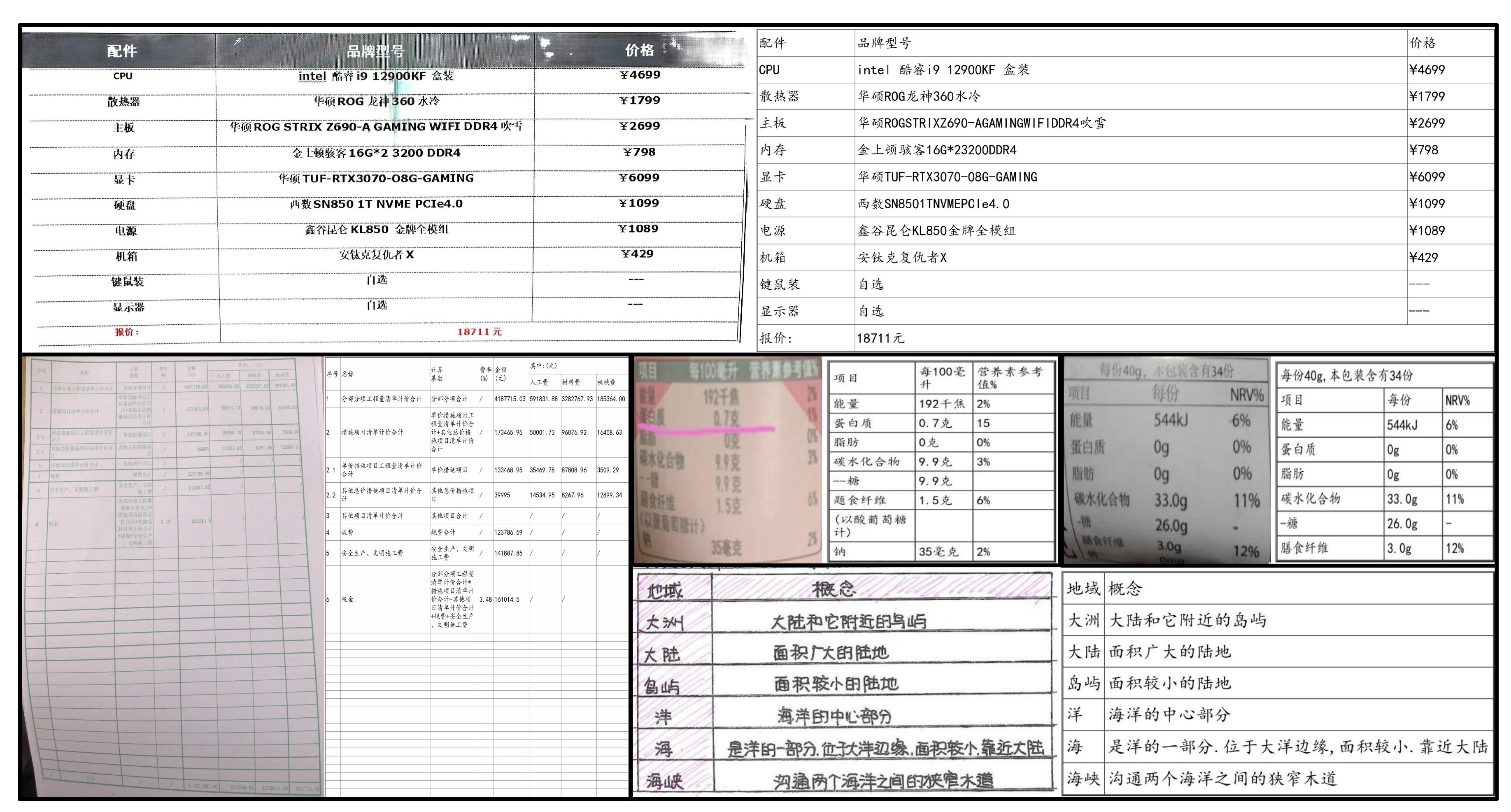}
    \caption{Visualization of table recognition results on real-world tables. In each subfigure, the left shows the input original table image, and the right presents the HTML-rendered visualization of the corresponding recognition result.}
    \label{fig:rw}
\end{figure*}

\begin{figure*}
    \centering
    \includegraphics[width=1\linewidth]{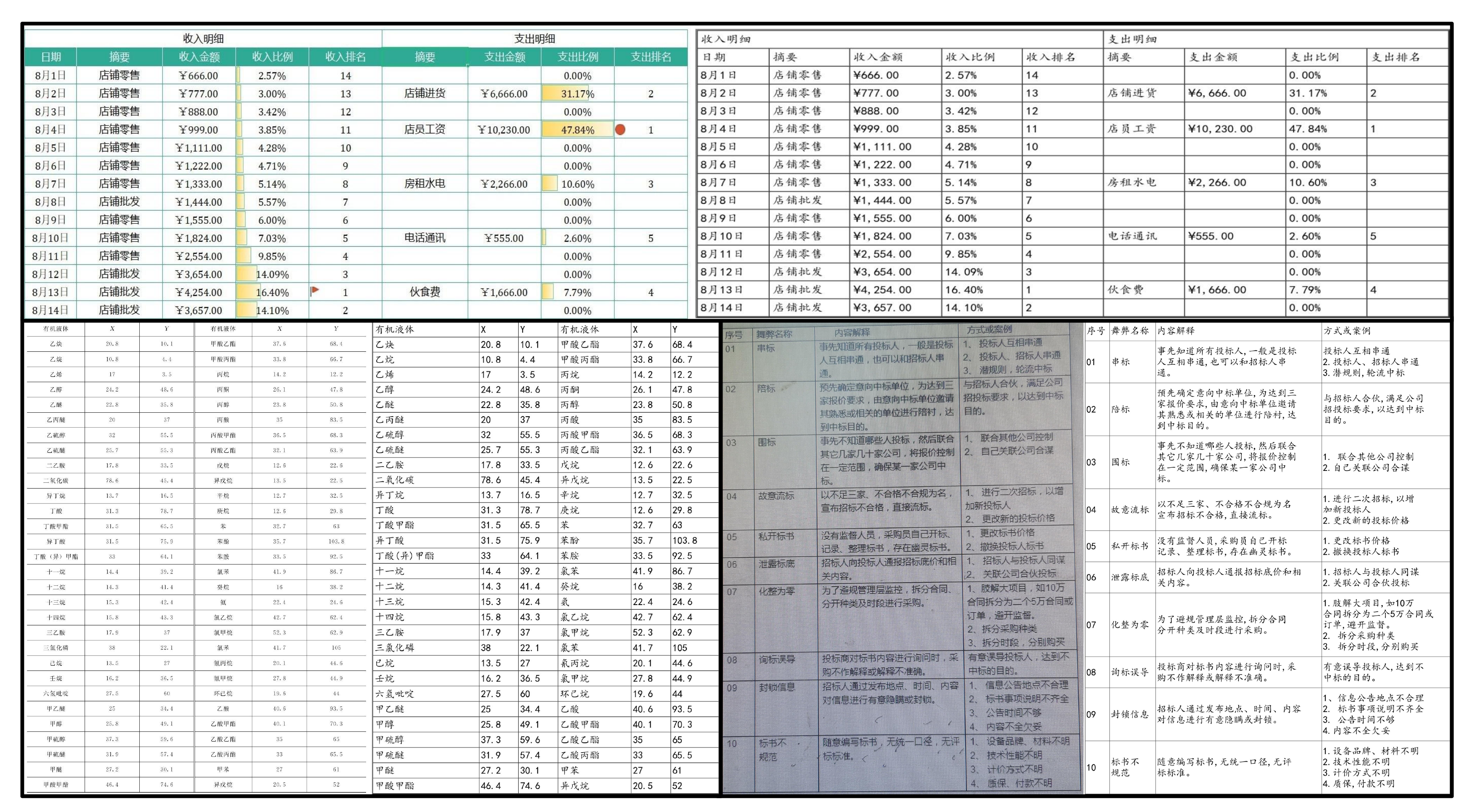}
    \caption{Visualization of table recognition results on long tables. In each subfigure, the left shows the input original table image, and the right presents the HTML-rendered visualization of the corresponding recognition result.}
    \label{fig:long}
\end{figure*}

\begin{figure*}
    \centering
    \includegraphics[width=1\linewidth]{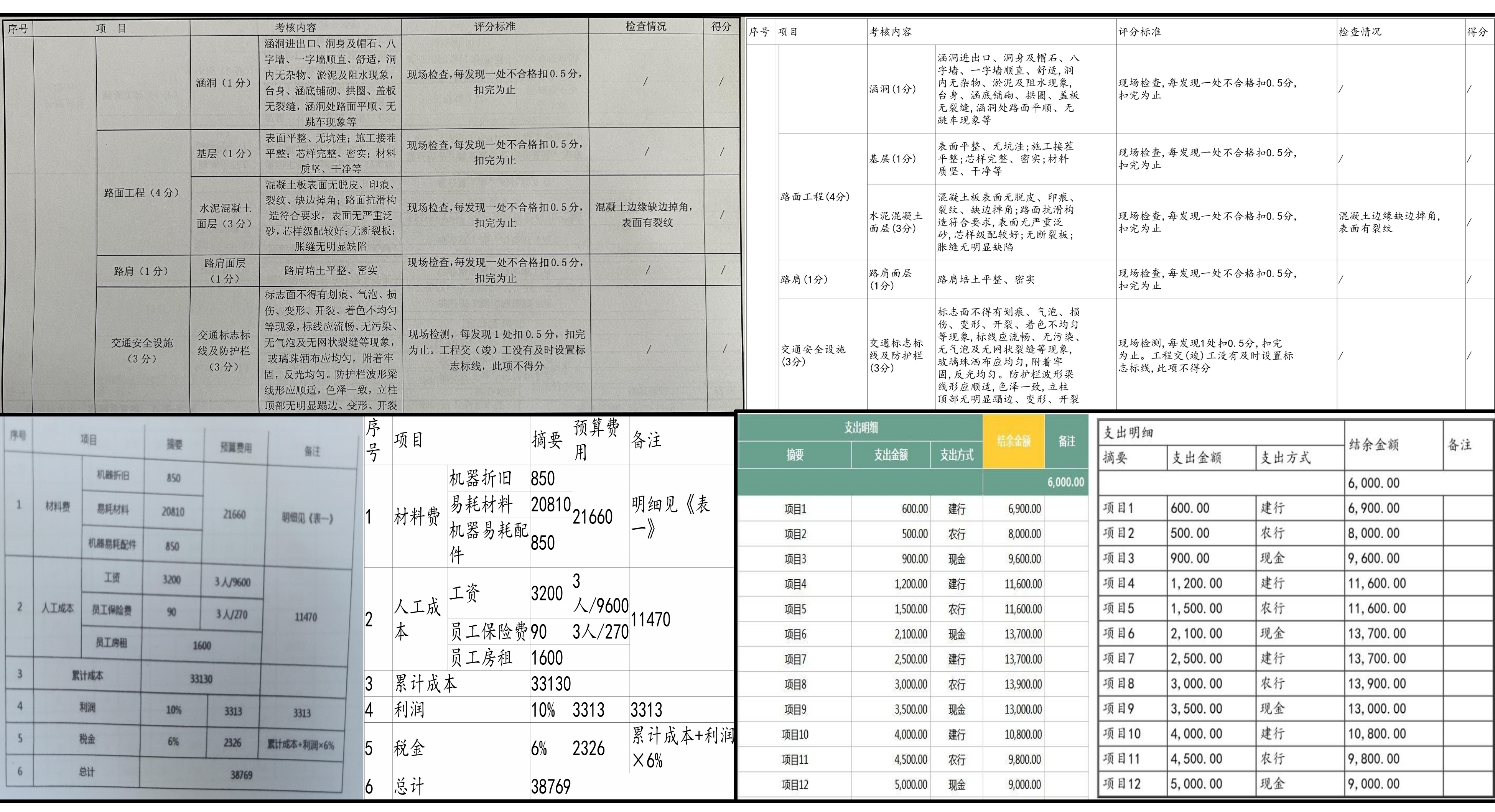}
    \caption{Visualization of table recognition results complex-structure tables. In each subfigure, the left shows the input original table image, and the right presents the HTML-rendered visualization of the corresponding recognition result.}
    \label{fig:comp}
\end{figure*}

We visualize several challenging table samples. Real-world tables (Fig.~\ref{fig:rw}) contain background noise, perspective distortion, and uneven illumination. Long tables (Fig.~\ref{fig:long}) feature lengthy sequences, numerous cells, and long text contents. Complex-structure tables (Fig.~\ref{fig:comp}) include extensive row or column spanning. Our method performs robustly across all these cases, demonstrating strong generalization and effectiveness.


\newpage
\section{Visualization of Cell Localization}
\label{sup:cell_vis}
\begin{figure*}
    \centering
    \includegraphics[width=1\linewidth]{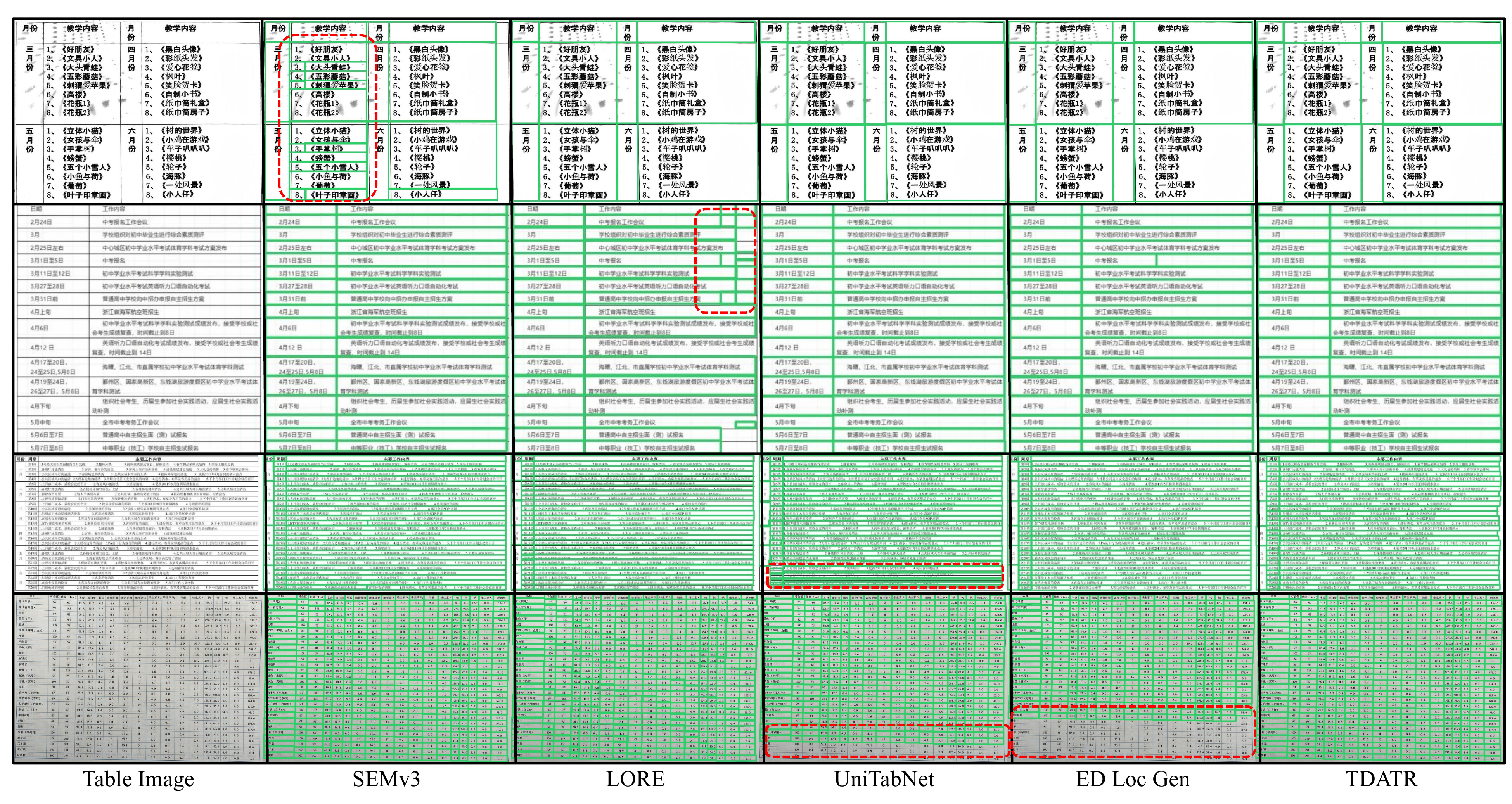}
    \caption{Qualitatively comparison of the cell localization.}
    \label{fig:cell_loc}
\end{figure*}

We qualitatively compare the cell localization results of several SOTA models,as shown in Fig.~\ref{fig:cell_loc}. SEMv3, which follows a ``split-and-merge'' strategy by detecting row/column separators to form table grids, is prone to confusing separators with inter-word gaps. LORE employs CornerNet for cell localization and relies solely on visual cues, making it unreliable for empty cells. UniTabNet predicts cell boxes through a single cell token, but compressing spatial information into one token limits its performance on dense tables. ``ED Loc Gen'' generates cell coordinates sequentially, resulting in excessively long answer sequences that are easily truncated on long tables.

\section{Failure Cases Analysis}
\label{failure_case}
We observed that these hard cases on iFLYTAB-full are mainly fall into three main error types:
(1) Boundary confusion: In borderless tables containing multi-line text the model struggles to distinguish text line spacing from cell delimiters.
(2) Span number errors: For cells with large row/column spans ( more than 15), the model occasionally predicts the error number. 
(3) Localization instability: Dense and empty borderless cells lack explicit visual cues causing instability visual-based regression in SGCL. 
We plan to enhance the decoder's semantic reasoning to resolve these visual ambiguities.

\end{document}